\documentclass[runningheads]{llncs}

\usepackage{amsmath}
\usepackage[table]{xcolor}
\usepackage{eccv}

\usepackage{eccvabbrv}

\usepackage{caption,subcaption}
\usepackage[T1]{fontenc}
\usepackage{array}
\usepackage{times}
\usepackage{epsfig}
\usepackage{graphicx}
\usepackage{amssymb}
\usepackage{physics}

\usepackage{epsfig}
\usepackage{epstopdf}
\usepackage{booktabs}
\usepackage{algpseudocode} 
\usepackage{lipsum}
\usepackage{multirow}
\usepackage{textcomp}
\usepackage{gensymb}
\usepackage{pifont}
\usepackage{bm}

\usepackage{tabularx}
\usepackage{adjustbox}
\usepackage{array}
\usepackage{multirow}
\usepackage{url}
\usepackage{booktabs}
\usepackage{array}
\usepackage{lipsum}
\usepackage{makecell} 
\usepackage{caption, subcaption}
\usepackage{enumitem}

\usepackage[accsupp]{axessibility}  

\newcommand{\cmark}{\ding{51}}%
\newcolumntype{R}[2]{%
    >{\adjustbox{angle=#1,lap=\width-(#2)}\bgroup}%
    l%
    <{\egroup}%
}

\newcommand\blfootnote[1]{%
  \begingroup
  \renewcommand\thefootnote{}\footnote{#1}%
  \addtocounter{footnote}{-1}%
  \endgroup
}

\usepackage{hyperref}

\usepackage{orcidlink}

\begin{document}

\title{CaesarNeRF: {Ca}librat{e}d {S}em{a}ntic {R}epresentation for Few-Shot Generalizable Neural Rendering} 

\titlerunning{{Ca}librat{e}d {S}em{a}ntic {R}epresentation for Few-Shot Generalizable Neural Rendering}

\author{Haidong Zhu$^{1,*}$\qquad \quad Tianyu Ding$^{2,*,\dagger}$\qquad \quad Tianyi Chen$^{2}$\qquad \quad Ilya Zharkov$^{2}$\qquad \quad \\Ram Nevatia$^{1}$\qquad \quad Luming Liang$^{2,\dagger}$
}

\authorrunning{H.~Zhu and T.~Ding et al.}

\institute{
$^{1}$University of Southern California\qquad \qquad $^{2}$Microsoft\\
\email{\tt\small\{haidongz,nevatia\}@usc.edu}\\
\email{\tt\small\{tianyuding,tiachen,zharkov,lulian\}@microsoft.com}}

\maketitle
\begin{center}
    \vspace{-.1in}
    \includegraphics[width=\linewidth]{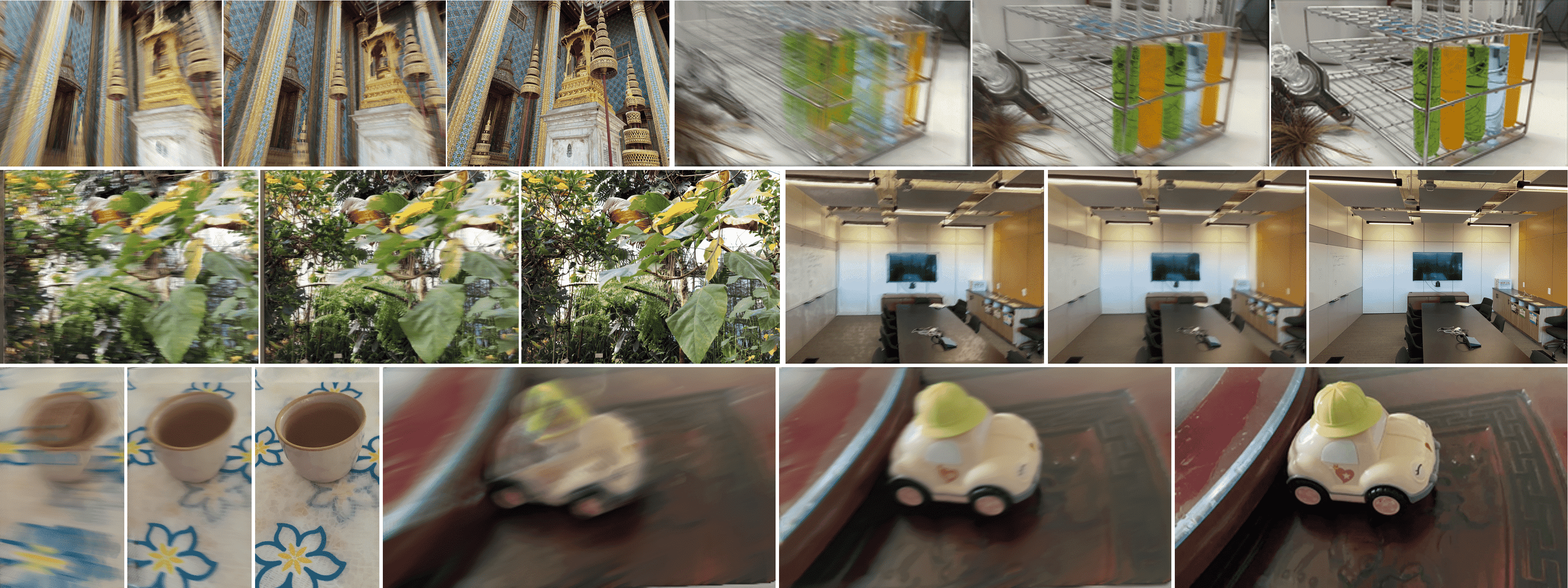}
    \captionof{figure}{Novel view synthesis for novel scenes using \textcolor{red}{\bf ONE} reference view on Shiny~\cite{wizadwongsa2021nex}, LLFF~\cite{mildenhall2019local}, and MVImgNet~\cite{yu2023mvimgnet} (top to bottom). Each triplet of images corresponds to the results from GNT~\cite{varma2022attention} (left), CaesarNeRF (middle) and groundtruth (right).}\label{fig:teaser}
  \end{center}

\begin{abstract}
  Generalizability and few-shot learning are key challenges in Neural Radiance Fields (NeRF), often due to the lack of a holistic understanding in pixel-level rendering. We introduce CaesarNeRF, an end-to-end approach that leverages scene-level \textbf{CA}librat\textbf{E}d \textbf{S}em\textbf{A}ntic \textbf{R}epresentation along with pixel-level representations to advance few-shot, generalizable neural rendering, facilitating a holistic understanding without compromising high-quality details. CaesarNeRF explicitly models pose differences of reference views to combine scene-level semantic representations, providing a calibrated holistic understanding. This calibration process aligns various viewpoints with precise location and is further enhanced by sequential refinement to capture varying details. Extensive experiments on public datasets, including LLFF, Shiny, mip-NeRF 360, and MVImgNet, show that CaesarNeRF delivers state-of-the-art performance across varying numbers of reference views, proving effective even with a single reference image. 
\blfootnote{$^*$Equal contribution.\qquad \qquad $^\dagger$Corresponding author.}
\blfootnote{\ \ This work was done when Haidong Zhu was an intern at  Microsoft.}

\end{abstract}

\section{Introduction}
Rendering a scene from a novel camera position is essential in view synthesis~\cite{buehler2001unstructured,debevec1996modeling,waechter2014let}. 
The recent advancement of Neural Radiance Field (NeRF)~\cite{mildenhall2021nerf} has shown impressive results in creating photo-realistic images from novel viewpoints. However, conventional NeRF methods are either typically scene-specific, necessitating retraining for novel scenes~\cite{mildenhall2021nerf,wizadwongsa2021nex,yang2023freenerf,fu2022multiplane,fridovich2023k}, or require a large number of reference views as input for generalizing to novel scenarios~\cite{yu2021pixelnerf,chen2021mvsnerf,suhail2022generalizable,varma2022attention}. These constraints highlight the complexity of the few-shot generalizable neural rendering, which aims to render unseen scenes from novel viewpoints with a limited number of reference images. 

Generalizing NeRF to novel scenes often involves using pixel-level feature embeddings encoded from input images, as seen in existing methods~\cite{yu2021pixelnerf,trevithick2021grf}. These methods adapt NeRF to novel scenes by separating the scene representation from the model through an image encoder. However, relying solely on pixel-level features has its drawbacks: it requires highly precise epipolar geometry and often overlooks occlusion in complex scenes. Moreover, employing pixel-level features ignores  the inherent interconnections within objects in the scene, treating the prediction of each pixel independently. Prior attempts to utilize  scene-level representation either suffers from style mismatches~\cite{liu2023zero} during scene-wide rendering or have been limited to specific object categories~\cite{jang2021codenerf,xie2021fig,mariotti2022viewnerf}. With few input reference under generalizable settings, these limitations become more pronounced, exacerbating the ambiguity in predictions due to biases from camera viewpoints. Although recent diffusion-based models~\cite{schwarz2020graf,gu2021stylenerf} attempt to address these issues through generative approaches, they struggle to effectively use the input views as contextual references for specific scenes.

We present CaesarNeRF, a method that advances the generalizability of NeRF by incorporating calibrated semantic representation. This enables rendering from novel viewpoints using as few as one input reference view, as depicted in Figure~\ref{fig:teaser}. Our approach combines semantic scene-level representation with per-pixel features, enhancing consistency across different views of the same scene. The encoder-generated scene-level representations capture both semantic features and biases linked to specific camera poses. When reference views are limited, these biases can introduce  uncertainty in the rendered images.
To counter this, CaesarNeRF integrates  camera pose transformations  into the semantic representation, {hence the term \emph{calibrated}}. 
By isolating  pose-specific information from the scene-level representation, our model harmonizes features across input views,
 mitigating view-specific biases and, in turn, reducing ambiguity.
In addition, CaesarNeRF introduces a sequential refinement process, which equips the model with varying levels of detail needed to enhance the semantic features. Extensive experiments on datasets such as LLFF~\cite{mildenhall2019local}, Shiny~\cite{wizadwongsa2021nex}, mip-NeRF 360~\cite{barron2022mip}, and the newly released MVImgNet~\cite{yu2023mvimgnet} demonstrate that CaesarNeRF outperforms current state-of-the-art methods with limited reference views available, proving effective  in generalizable settings with as few as one reference view. The project can be found \href{https://haidongz-usc.github.io/project/caesarnerf}{here}.

In summary, our contributions are as follows:
\begin{itemize}
    \item We introduce CaesarNeRF, which utilizes scene-level calibrated semantic representation to achieve few-shot, generalizable neural rendering. This innovation leads to coherent and high-quality renderings.
    
    \item  We integrate  semantic scene context with pixel-level details, in contrast to existing methods that rely solely on pixel-level features. We also address view-specific biases by modeling camera pose transformations and enhance the scene understanding through the sequential refinement of semantic features.
    
    \item We demonstrate through extensive experiments that CaesarNeRF consistently outperforms state-of-the-art generalizable NeRF methods across a variety of datasets. Furthermore, integrating the Caesar pipeline into other baseline methods leads to consistent performance gains, highlighting its effectiveness and adaptability.
\end{itemize}

\section{Related work}

\textbf{Neural Radiance Field (NeRF)} implicitly captures the density and appearance of points within a scene or object~\cite{mildenhall2021nerf,mildenhall2020nerf} and enables rendering from novel camera positions. In recent years, NeRF has witnessed improvements in a wide range of applications, such as photo-realistic novel view synthesis for large-scale scenes~\cite{martin2021nerf,xiangli2022bungeenerf,zhenxing2022switch}, dynamic scene decomposition and deformation~\cite{zhuang2022mofanerf,liu2021neural,noguchi2021neural,park2021nerfies,peng2021animatable,pumarola2021d,zhu2023cat,jiang2023alignerf,li2023dynibar}, occupancy or depth estimation~\cite{xu2022point,wei2021nerfingmvs,zhu2023multimodal}, scene generation and editing~\cite{zhang2023text2nerf,metzer2023latent,poole2022dreamfusion,lin2023magic3d,jain2022zero,yang2021learning,lin2023componerf,yang2023contranerf,bao2023sine}, and so on. Despite these advances, most methods still rely on the original NeRF and require retraining or fine-tuning for novel scenes not covered in the training data.

\textbf{Generalizable NeRF} aims to adapt a single NeRF model to multiple scenes by separating the scene representation from the model. This field has seen notable advancements, with efforts focused on avoiding the need for retraining~\cite{irshad2023neo,yu2021pixelnerf,yao2018mvsnet,wang2021ibrnet,wang2023rodin,liu2022neural,chen2023explicit}. PixelNeRF~\cite{yu2021pixelnerf} and GRF~\cite{trevithick2021grf} pioneered the application of an image encoder to transform images into per-pixel features, with NeRF functioning as a decoder for predicting density and color from these features. MVSNeRF~\cite{chen2021mvsnerf} introduces the use of a cost volume from MVSNet~\cite{yao2018mvsnet} to encode 3-D features from multiple views. Recognizing the intrinsic connection between points along a ray, IBRNet~\cite{wang2021ibrnet} employs self-attention to enhance point density predictions. 
Transformer-based~\cite{vaswani2017attention} networks like GNT~\cite{varma2022attention,cong2023enhancing}, GeoNeRF~\cite{johari2022geonerf}, and GPNR~\cite{suhail2022generalizable} are explored as alternatives to volume rendering, concentrating on pixel and patch-level representations. Additionally, InsertNeRF~\cite{bao2023insertnerf} utilizes hypernet modules to adapt parameters for novel scenes efficiently.

These methods primarily depend on image encoders to extract pixel-aligned features from reference views. As a result, many of them lack a comprehensive understanding of the entire scene. Furthermore, with few reference views, the features become intertwined with view-specific details, compromising the quality of the rendering results.

\textbf{Few-shot Neural Radiance Field} aims to render novel views using a limited number of reference images. To this end, various methods have been developed, incorporating additional information such as normalization-flow~\cite{zhang2021ners}, semantic constraints~\cite{jain2021putting,gao2023surfelnerf}, depth cues~\cite{roessle2022dense,deng2022depth}, geometry consistency~\cite{kwak2023geconerf,niemeyer2022regnerf,wang2023sparsenerf,bao2023and,xu2022sinnerf}, and frequency content~\cite{yang2023freenerf}. Others~\cite{chen2021mvsnerf,chibane2021stereo} emphasize pretraining on large-scale datasets.

While these methods offer reasonable reconstructions with few inputs, they typically still require training or fine-tuning for specific scenes. Moreover, these methods usually require at least three reference images. With fewer than three, view-specific biases lead to ambiguity, complicating the rendering. Diffusion-based~\cite{liu2023zero,Deng_2023_CVPR,zhou2023sparsefusion,shue20233d} and other generative methods~\cite{kania2023hypernerfgan,zimny2022points2nerf} have been explored for single-view synthesis or generative rendering, yet they are mostly limited to single-object rendering and generally fall short for complex scenes, which often result in a style change.

CaesarNeRF confronts the above challenges by leveraging calibrated semantic representations  that exploit scene geometry and variations in camera viewpoints. 
As a result, CaesarNeRF overcomes the limitations of pixel-level features and reduces dependency on external data or extensive pretraining, delivering high-quality renderings in few-shot and generalizable settings.

\section{The proposed method}

\begin{figure*}[t]
    \centering
    \includegraphics[width=0.9\linewidth]{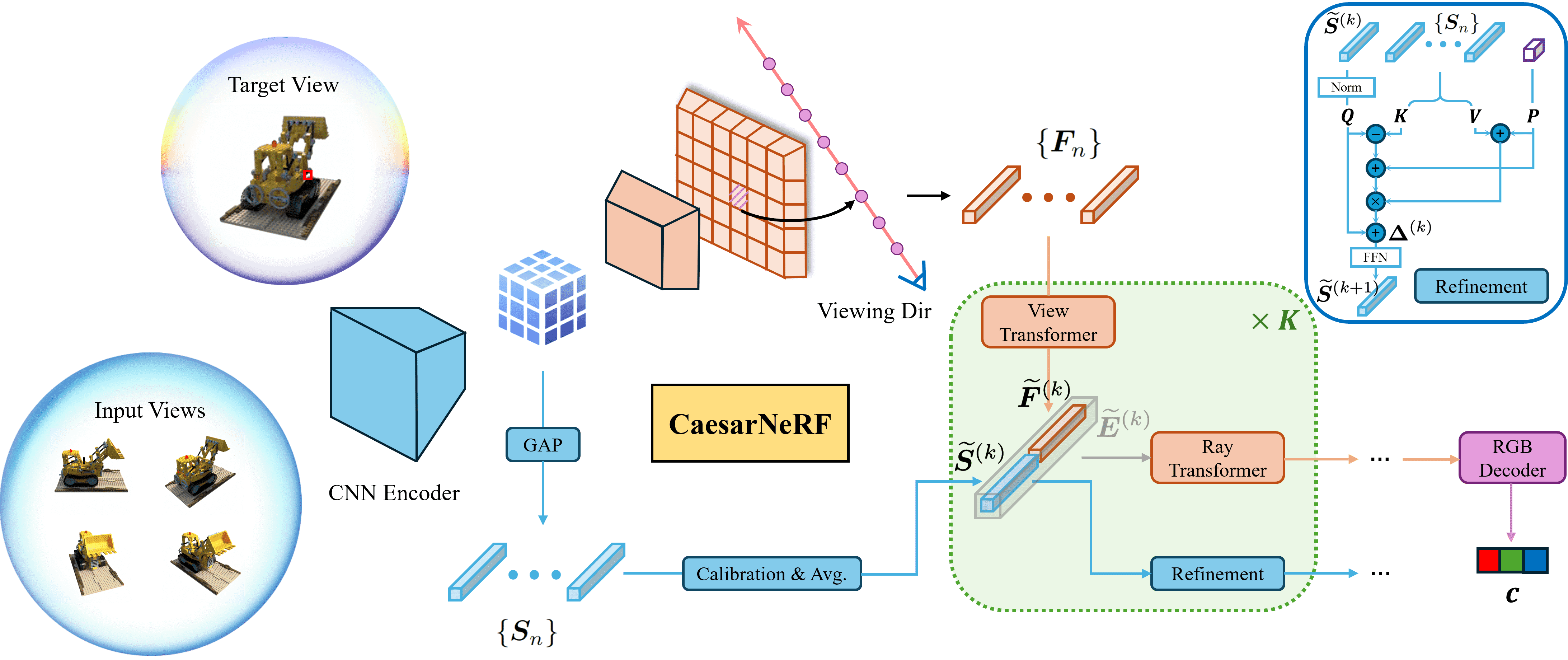}
    \caption{
    Overview of CaesarNeRF. CaesarNeRF employs a shared encoder to capture two types of features from input views, including scene-level semantic representation $\{\bm S_n\}$ and pixel-level feature representation $\{\bm F_n\}$. We use the same encoder for both the scene-level semantic representation and the pixel-level embeddings. Following calibration and aggregation of $\{\bm S_n\}$ from various views, we concatenate it with the pixel-level fused feature, processed by the view transformer. Subsequent use of the ray-transformer, coupled with sequential refinement, enables us to render the final RGB values for each pixel in the target view. The output features serve as the input for the next stage, indicated by matching line colors.
    }
    \label{fig:pipeline}
\end{figure*}

We first outline the general framework of existing generalizable NeRF in Section~\ref{sec:gnrf}. Then, we present our proposed CaesarNeRF, as illustrated in Figure~\ref{fig:pipeline}. This model integrates elements of semantic representation, calibration, and sequential refinement, detailed in Section~\ref{sec:method1}, \ref{sec:cali}, and \ref{sec:seq}, respectively. The training objective is given in~\ref{sec:obj}.

\subsection{NeRF and generalizable NeRF}\label{sec:gnrf}

Neural Radiance Field (NeRF)~\cite{mildenhall2021nerf, mildenhall2020nerf} aims to render 3D scenes by predicting both the density and RGB values at points where light rays intersect the radiance field. For a query point $\bm x\in\mathbb{R}^3$ and a viewing direction $\bm d$ on the unit sphere $\mathbb{S}^2$ in 3D space, the NeRF model $\bm{\mathcal{F}}$ is defined as:
\begin{align}
    \sigma, \bm c = \bm{\mathcal{F}}(\bm x, \bm d).
\end{align}
Here, $\sigma\in\mathbb{R}$ and $\bm c\in\mathbb{R}^3$ denote the density and the RGB values, respectively. After computing these values for a collection of  discretized points along each ray, volume rendering techniques are employed to calculate the final RGB values for each pixel, thus reconstructing the image.

However, traditional NeRF models $\bm{\mathcal{F}}$ are limited by their requirement for scene-specific training, making it unsuitable for generalizing to novel scenes. To overcome this, generalizable NeRF models, denoted by $\bm{\mathcal{F}_G}$, are designed to render images of novel scenes without per-scene training. Given $N$ reference images $\{\vb*I_n\}_{n=1}^{N}$, an encoder-based generalizable NeRF 
model $\bm{\mathcal{F}_G}$ decouples the object representation from the original NeRF by using an encoder to extract per-pixel feature maps $\{\vb*F_n\}_{n=1}^{N}$ from the input images. To synthesize a pixel associated with a point $\bm x$ along a ray in direction $\bm d$, it projects $\{\vb*F_n\}_{n=1}^{N}$ from nearby views and aggregates this multi-view pixel-level information using techniques such as average pooling~\cite{yu2021pixelnerf} or cost volumes~\cite{chen2021mvsnerf}. This results in a fused feature embedding $\widetilde{\bm F}$, allowing $\bm{\mathcal{F}_G}$ to predict density $\sigma$ and RGB values $\bm c$ for each point along the ray, as expressed by:
\begin{align}\label{eq:gnerf}
    \sigma, \bm c = \bm{\mathcal{F}_G}(\bm x, \bm d, \widetilde{\bm F}).
\end{align}

In our method, we adopt the recently introduced fully attention-based generalizable NeRF method, GNT~\cite{varma2022attention}, as both the backbone and the baseline. GNT shares a similar paradigm with~\eqref{eq:gnerf} but employs transformers~\cite{vaswani2017attention} to aggregate pixel-level features into $\widetilde{\bm F}$. It uses a \textit{view transformer} to fuse projected pixel-level features from reference views, and a \textit{ray transformer} to combine features from different points along a ray, eliminating the need for volume rendering. Further details about GNT can be found in~\cite{varma2022attention}. We also demonstrate that our approach can be extended to other generalizable NeRF models, as discusses in Section~\ref{sec:results}.

\subsection{Scene-level semantic representation}\label{sec:method1}

Both encoder-based generalizable NeRF models~\cite{yu2021pixelnerf,chen2021mvsnerf,trevithick2021grf} and their attention-based counterparts~\cite{varma2022attention,suhail2022generalizable} mainly rely on pixel-level feature representations. While effective, this approach restricts their capability for a holistic scene understanding, especially when reference views are scarce. This limitation also exacerbates challenges in resolving depth ambiguities between points along the rays, a problem that becomes more pronounced with fewer reference views.

To address these challenges, we introduce semantic representations aimed at enriching the scene-level understanding. We utilize a shared CNN encoder and apply a Global Average Pooling (GAP) to its $C$-dimensional output feature map, generating $N$ \emph{global} feature vectors $\{\vb*S_n\}_{n=1}^N$ corresponding to each input view. These feature vectors are then averaged to form a unified scene-level representation $\vb*S$, \ie,
\begin{align}\label{eq:scene}
    \vb*S=\frac{1}{N}\sum_{n=1}^N \bm S_n \in\mathbb{R}^C.
\end{align}
In GNT~\cite{varma2022attention}, which uses a view transformer to aggregate pixel-level features into an $L$-dimensional vector $\widetilde{\bm F}$, we extend this by concatenating $\widetilde{\bm F}$ with $\vb*S$ to construct a \emph{global-local} embedding $\vb*E$, as formulated by:
\begin{align}\label{eq:concat}
    \vb*E = \text{Concat}(\widetilde{\bm F}, \vb*S)\in\mathbb{R}^{L+C}.
\end{align}
This combined embedding $\vb*E$ is then subjected to the standard self-attention mechanism used in GNT~\cite{varma2022attention}. This approach enables the scene-level semantic representation ($\bm S$) to integrate with per-point features ($\widetilde{\vb*F}$), offering a more nuanced understanding at both levels. It also allows each point to selectively draw from the scene-level information. To maintain dimensional consistency across the input and output layers of multiple transformer modules, we employ a two-layer MLP to project the enhanced features back to the original dimension $L$ of the per-point embedding $\widetilde{\bm F}$.

\begin{figure}[t]
    \centering
    \resizebox{0.7\textwidth}{!}
    {
    \begin{tabularx}{\textwidth}{cc}
        \includegraphics[width=0.45\linewidth]{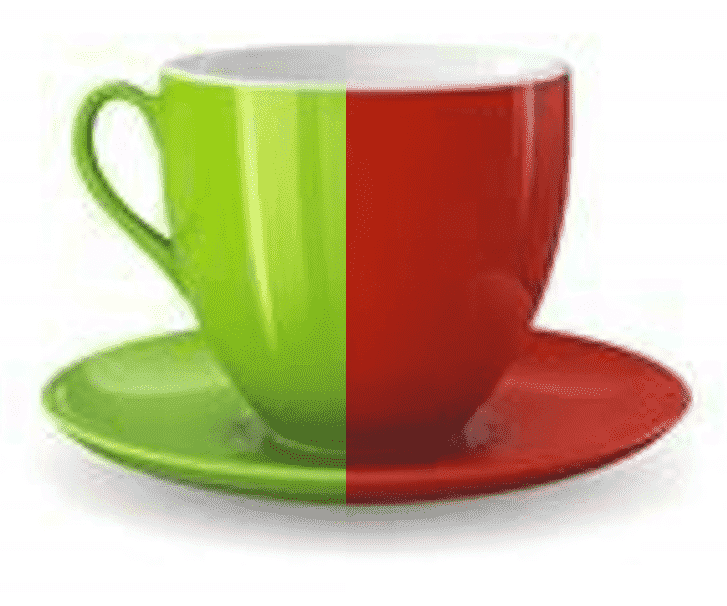} &  \includegraphics[width=0.45\linewidth]{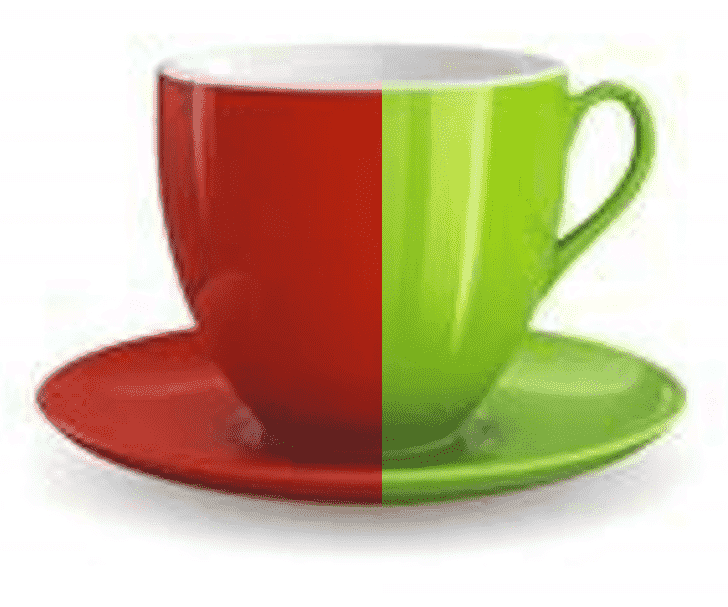} \\
        (a) ``L: green, R: red" & 
        (b) ``L: red, R: green"\\
    \end{tabularx}
    }
    \caption{
     An illustration of conflicting semantic meanings from multiple viewpoints of the same object. When observing the cup from distinct angles, the features extracted after pooling retain spatial information but are inconsistent in the scene-level semantic understanding, leading to conflicts across various reference images after aggregation.}
    \label{fig:conflict}
\end{figure}

\subsection{Calibration of semantic representation}\label{sec:cali}

The integration of the scene-level semantic representation $\vb*S$, generated through simple averaging of global feature vectors as in~\eqref{eq:scene}, improves rendering quality. However, this approach has limitations when dealing with multiple views. As illustrated in Figure~\ref{fig:conflict}, viewing the same object from distinct angles may retain spatial attributes but can lead to conflicting semantic meanings. Merely averaging these global feature vectors without accounting for camera positions can result in a distorted scene-level understanding.

To mitigate this inconsistency, we propose a semantic calibration technique using feature rotation. This adjustment aligns the semantic representation across different camera poses. Our inspiration comes from the use of camera pose projection in computing the fused pixel-level feature $\widetilde{\bm F}$ and is further motivated by~\cite{qi2017pointnet}, which demonstrates that explicit rotation operations in feature spaces are feasible. Unlike point clouds in~\cite{qi2017pointnet} that inherently lack a defined canonical orientation, NeRF explicitly encodes differences between camera viewpoints, thereby enabling precise calibration between the reference and target images.

Building on this observation, we calculate calibrated semantic representations $\{\widetilde{\vb*S}_n\}_{n=1}^N$ from the $N$ original semantic representations $\{\vb*S_n\}_{n=1}^N$ derived from the reference views. We accomplish this by leveraging their respective rotation matrices $\{\vb*T_n\}_{n=1}^N$ to model the rotational variations between each input view and the target view. The alignment of the original semantic features is performed as follows:
\begin{align}
\begin{split}
    \widetilde{\vb*S}_n &= \bm{\mathcal{P}}( \vb*T_n \cdot \bm{\mathcal{P}}^{-1}(\vb*S_n)), \ \text{ where } \vb*T_n= \vb*T^{\text{w2c}}_{\text{out}} \cdot \vb*T^{\text{c2w}}_n.
\end{split}
\end{align}
Here, $\vb*T^{\text{c2w}}_n$ is the inverse of the extrinsic matrix used for $\vb*I_n$, and $\vb*T^{\text{w2c}}_{\text{out}}$ is the extrinsic matrix for the target view. $\bm{\mathcal{P}}(\cdot)$ and $\bm{\mathcal{P}}^{-1}(\cdot)$ are the flattening and inverse flattening operations, which reshape the feature to a 1D vector of shape $1$-by-$C$ and a 2D matrix of shape $3$-by-$\frac{C}{3}$, respectively.

Note that for the extrinsic matrix, we consider only the top-left $3 \times 3$ submatrix that accounts for rotation. 
Using GAP to condense feature maps of various sizes into a $1$-by-$C$ feature vector eliminates the need for scaling parameters in the semantic representation. As a result, modeling the intrinsic matrix is unnecessary, assuming no skewing, making our approach adaptable to different camera configurations.

With the calibrated semantic features $\{\widetilde{\vb*S}_n\}_{n=1}^N$ for each reference view, we average these, similar to~\eqref{eq:scene}, to obtain the calibrated scene-level semantic representation $\widetilde{\vb*S}$, \ie,
\begin{equation}
   \widetilde{\vb*S}=\frac{1}{N}\sum_{n=1}^N \widetilde{\bm S}_n\in\mathbb{R}^{C}.
\end{equation}
Finally, akin to~\eqref{eq:concat}, we concatenate the pixel-level fused feature $\widetilde{\bm F}$ with the calibrated scene-level semantic representation $\widetilde{\vb*S}$ to form the final global-local embedding $\widetilde{\vb*E}$:
\begin{align}\label{eq:concat2}
    \widetilde{\vb*E} = \text{Concat}(\widetilde{\bm F}, \widetilde{\vb*S})\in\mathbb{R}^{L+C}.
\end{align}
This unified embedding then feeds into ray transformers, passing through standard self-attention mechanisms. In the original GNT~\cite{varma2022attention}, multiple view transformers and ray transformers are stacked alternately for sequential feature processing. The last ray transformer integrates features from multiple points along a ray to yield the final RGB value. We denote the corresponding feature representations at stage $k$ as $\widetilde{\bm F}^{(k)}$ and $\widetilde{\bm E}^{(k)}$. Notably, the calibrated semantic representation $\widetilde{\bm S}$ remains constant across these stages.

\begin{figure}[t]
    \centering
    \includegraphics[width=\linewidth]{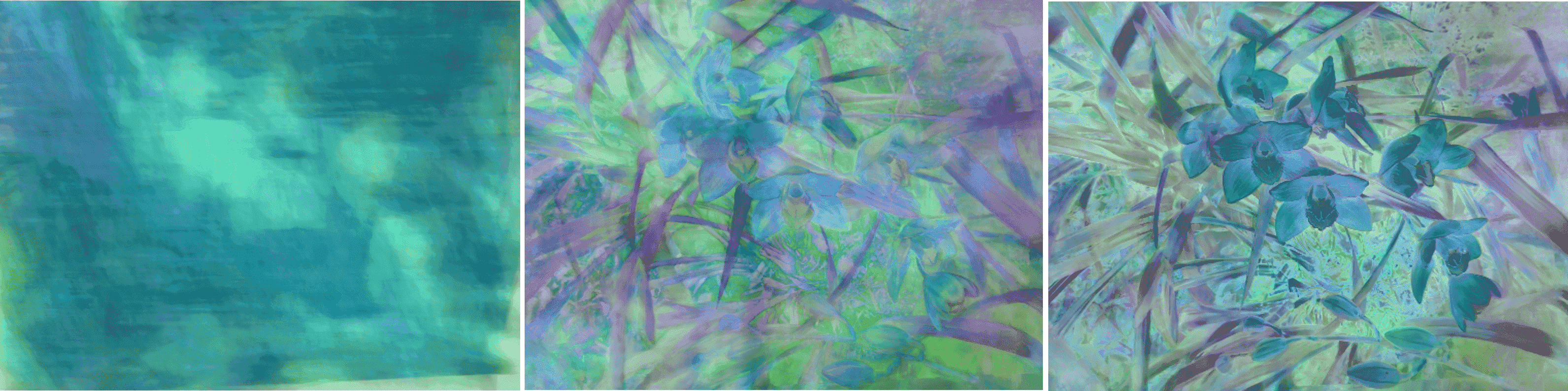}
    \caption{
    Visualization of decoded feature maps for ``orchid'' in LLFF dataset, produced by \textit{ray transformers}~\cite{varma2022attention} at different stages. From left to right, the transformer stages increase in depth.}
    \label{fig:seqdec}
\end{figure}

\subsection{Sequential refinement}\label{sec:seq}

While leveraging $\widetilde{\bm S}$ improves consistency, a single, uniform $\widetilde{\bm S}$ may not be adequate for deeper layers that demand more nuanced details. In fact, we find that deeper transformers capture finer details compared to shallower ones, as shown in Figure~\ref{fig:seqdec}. To address this limitation, we introduce a sequential semantic feature refinement module that progressively enriches features at each stage. Specifically,  we learn the residual $\bm\Delta^{(k)}$ to update $\widetilde{\vb*S}$ at each stage $k$ as follows:
\begin{align}
    \widetilde{\vb*S}^{(k+1)} &\leftarrow \widetilde{\vb*S}^{(k)} + \bm\Delta^{(k)}.
\end{align}
Here, $\bm\Delta^{(k)}$ is calculated by first performing specialized cross-attentions between $\widetilde{\bm S}^{(k)}$ and the original, uncalibrated per-frame semantic features $\{\bm S_n\}_{n=1}^N$ (see Figure~\ref{fig:pipeline}), followed by their summation. Our goal is to fuse information from different source views to enrich the scene-level semantic representation with features from each reference frame.
With this sequential refinement, we combine $\widetilde{\bm S}^{(k)}$ with $\widetilde{\bm F}^{(k)}$ at each stage, yielding a stage-specific global-local embedding $\widetilde{\bm E}^{(k)}$, which completes our approach.

\textbf{Discussion.} In scenarios with few reference views, especially when limited to just one, the primary issue is inaccurate depth estimation, resulting in depth ambiguity~\cite{Deng_2023_CVPR}. This compromises the quality of images when rendered from novel viewpoints. Despite this, essential visual information generally remains accurate across different camera poses. Incorporating our proposed scene-level representation improves the understanding of the overall scene layout~\cite{chen2023beyond}, distinguishing our approach from existing generalizable NeRF models that predict pixels individually. The advantage of our approach is its holistic view; the semantic representation enriches per-pixel predictions by providing broader context.  This semantic constraint ensures that  fewer abrupt changes between adjacent points. Consequently, it leads to more reliable depth estimations, making the images rendered from limited reference views more plausible.

\subsection{Training objectives}\label{sec:obj}

During training, we employ three different loss functions:

\textbf{MSE loss.} The Mean Square Error (MSE) loss is the standard photometric loss used in NeRF~\cite{mildenhall2020nerf}. It computes the MSE between the actual and predicted pixel values.

\textbf{Central loss.} Since we project the view-level features from different input camera poses to the shared target view, we introduce a central loss to ensure frame-wise calibrated semantic features $\{\widetilde{\bm S}_n\}_{n=1}^N$ are consistent when projected onto the same target view, which is defined as:
\begin{align}
    \mathcal{L}_{\text{central}}= \frac{1}{N}\sum_{n=1}^N\left\| \widetilde{\vb*S}_n - \widetilde{\vb*S}\right\|_1.
\end{align}

\textbf{Point-wise perceptual loss.} During the rendering of a batch of pixels in a target view, we inpaint the ground-truth image by replacing the corresponding pixels with the predicted ones. Then, a perceptual loss~\cite{johnson2016perceptual} is computed between the inpainted image and the target image to guide the training process at the whole-image level.

The final loss function is formulated as follows:
\begin{equation}
    \mathcal{L} = \mathcal{L}_{\text{MSE}} + \lambda_1 \mathcal{L}_{\text{central}} + \lambda_2 \mathcal{L}_{\text{perc}}.
\end{equation}
Empirically, we set $\lambda_1=1$ and $\lambda_2=0.001$, following~\cite{lin2022efficient}.

\section{Experiments}

In this section, we first discuss our experimental setups and other details required for our experiments in~\Cref{sec:exp}, followed by our results for different experimental settings, along with ablation studies and analysis in~\Cref{sec:results}.

\subsection{Experimental setups}\label{sec:exp} 

\textbf{Datasets.}  Firstly, following~\cite{varma2022attention}, we construct our training data from both synthetic and real data. This collection includes scanned models from Google Scanned Objects~\cite{downs2022google}, RealEstate10K~\cite{zhou2018stereo}, and handheld phone captures~\cite{wang2021ibrnet}. For evaluation, we utilize real data encompassing complex scenes from sources such as LLFF~\cite{mildenhall2019local}, Shiny~\cite{wizadwongsa2021nex}, and mip-NeRF 360~\cite{barron2022mip}. Additionally, we train and test our model using the recently released MVImgNet dataset~\cite{yu2023mvimgnet}. We adhere to the official split, focusing on examples from the \textit{containers} category, and select 2,500 scenes for training. During inference, we choose 100 scenes, using their first images as target views and the spatially nearest images as references. Since MVImgNet does not provide camera poses, we utilize COLMAP~\cite{schoenberger2016sfm,schoenberger2016mvs} to deduce the camera positions within these scenes.

\begin{table}[tb]
\centering
\caption{Results for generalizable scene rendering on LLFF with few reference views. GeoNeRF, MatchNeRF, and MVSNeRF necessitate variance as input, defaulting to 0 for single-image cases, hence their results are not included for 1-view scenarios.} 
\label{tab:main_few_llff}
\def\lw{1.5}
\def\ls{0.01}
\resizebox{\linewidth}{!}
{
\begin{tabular}{p{2.3cm}p{\ls cm}p{\lw cm}<{\centering}p{\lw cm}<{\centering}p{\lw cm}<{\centering}p{\ls cm}p{\lw cm}<{\centering}p{\lw cm}<{\centering}p{\lw cm}<{\centering}p{\ls cm}p{\lw cm}<{\centering}p{\lw cm}<{\centering}p{\lw cm}<{\centering}} \toprule
\multirow{2}{*}{Method} &&\multicolumn{3}{c}{1 reference view} && \multicolumn{3}{c}{2 reference views} && \multicolumn{3}{c}{3 reference views}\\
\cline{3-5} \cline{7-9} \cline{11-13} \\ [-8pt]
&&PSNR ($\uparrow$) & LPIPS ($\downarrow$) & SSIM ($\uparrow$) && PSNR ($\uparrow$) & LPIPS ($\downarrow$) & SSIM ($\uparrow$) && PSNR ($\uparrow$) & LPIPS ($\downarrow$) & SSIM ($\uparrow$)\\
\midrule
PixelNeRF \cite{yu2021pixelnerf} && 9.32 & 0.898 & 0.264 && 11.23 & 0.766 & 0.282 && 11.24 & 0.671 & 0.486 \\
GPNR \cite{suhail2022generalizable} && 15.91 & \cellcolor{yellow!40}0.527 & 0.400 && 18.79 & 0.380 & 0.575  && 21.57 & 0.288 & 0.695 \\
NeuRay \cite{liu2022neural}  && 16.18 & 0.584 & 0.393 && 17.71 & 0.336 & 0.646 && 18.26 & 0.310 & 0.672 \\
GeoNeRF \cite{johari2022geonerf} && - & - & - && 18.76 & 0.473 & 0.500 && \cellcolor{orange!40}23.40 & 0.246 & \cellcolor{yellow!40}0.766\\
MatchNeRF \cite{chen2023explicit} && - & - & - && \cellcolor{yellow!40}21.08 & \cellcolor{yellow!40}0.272 & 0.689 && 22.30 & \cellcolor{yellow!40}0.234 & 0.731 \\
MVSNeRF \cite{chen2021mvsnerf} && - & - & - && 19.15 & 0.336& \cellcolor{orange!40}0.704 && 19.84 & 0.314 & 0.729 \\
IBRNet \cite{wang2021ibrnet} && \cellcolor{orange!40}16.85 & 0.542 & \cellcolor{orange!40}0.507 && \cellcolor{orange!40}21.25 & 0.333 & 0.685 && 23.00 & 0.262 & 0.752 \\
GNT \cite{varma2022attention} && \cellcolor{yellow!40}16.57 & \cellcolor{orange!40}0.500& \cellcolor{yellow!40}0.424 && 20.88 & \cellcolor{orange!40}0.251 & \cellcolor{yellow!40}0.691 && \cellcolor{yellow!40}23.21 & \cellcolor{orange!40}0.178 & \cellcolor{orange!40}0.782 \\ 
Ours && \cellcolor{red!40}18.31 & \cellcolor{red!40}0.435 & \cellcolor{red!40}0.521 && \cellcolor{red!40}21.94 & \cellcolor{red!40}0.224 & \cellcolor{red!40}0.736 && \cellcolor{red!40}23.45 & \cellcolor{red!40}0.176 & \cellcolor{red!40}0.794 \\ 
\bottomrule
\end{tabular}
}
\end{table}

\textbf{Implementation details.} CaesarNeRF is built upon GNT~\cite{varma2022attention}, for which we maintain the same configuration, setting the \textit{ray} and \textit{view} transformers stack number ($K$) to 8 for generalizable setting and 4 for single-scene setting. The feature encoder  extracts bottleneck features, applies GAP,  and then uses a fully connected (FC) layer to reduce the input dimension $C$ to 96. Training involves 500,000 iterations using the Adam optimizer~\cite{kingma2014adam}, with learning rate set at 0.001 for the feature encoder and 0.0005 for CaesarNeRF, halving them every 100,000 iterations. Each iteration  samples 4,096 rays from a single scene. In line with~\cite{varma2022attention}, we randomly choose between 8 to 10 reference views for training, and 3 to 7 views when using the MVImgNet~\cite{yu2023mvimgnet}.

\textbf{Baseline methods.} We compare CaesarNeRF with several state-of-the-art methods suited for generalizable NeRF applications, including earlier works such as  MVSNeRF~\cite{chen2021mvsnerf}, PixelNeRF~\cite{yu2021pixelnerf}, and IBRNet~\cite{wang2021ibrnet}, alongside more recent ones, including GPNR~\cite{suhail2022generalizable}, NeuRay~\cite{liu2022neural}, GNT~\cite{varma2022attention}, GeoNeRF~\cite{johari2022geonerf} and MatchNeRF~\cite{chen2023explicit}.

\subsection{Results and analysis}\label{sec:results}

We compare results in two settings: a generalizable setting, where the model is trained on multiple scenes without fine-tuning during inference for both few and all reference view cases, and a single-scene setting where the model is trained and evaluated on just one scene. Following these comparisons, we conduct ablation studies and test the generalizability of our method with other state-of-the-art approaches.

\begin{table}[tb]
\centering
\caption{Results for generalizable scene rendering on Shiny with few reference views.} 
\label{tab:main_shiny}
\def\lw{1.5}
\def\ls{0.01}
\resizebox{\linewidth}{!}
{
\begin{tabular}{p{2.3cm}p{\ls cm}p{\lw cm}<{\centering}p{\lw cm}<{\centering}p{\lw cm}<{\centering}p{\ls cm}p{\lw cm}<{\centering}p{\lw cm}<{\centering}p{\lw cm}<{\centering}p{\ls cm}p{\lw cm}<{\centering}p{\lw cm}<{\centering}p{\lw cm}<{\centering}} \toprule
\multirow{2}{*}{Method} &&\multicolumn{3}{c}{1 reference view} && \multicolumn{3}{c}{2 reference views} && \multicolumn{3}{c}{3 reference views}\\
\cline{3-5} \cline{7-9} \cline{11-13} \\ [-8pt]
&&PSNR ($\uparrow$) & LPIPS ($\downarrow$) & SSIM ($\uparrow$) && PSNR ($\uparrow$) & LPIPS ($\downarrow$) & SSIM ($\uparrow$) && PSNR ($\uparrow$) & LPIPS ($\downarrow$) & SSIM ($\uparrow$)\\
\midrule 
MatchNeRF \cite{chen2023explicit} && - & - & - && \cellcolor{yellow!40}20.28 & \cellcolor{red!40}0.278 & \cellcolor{orange!40}0.636 &&  20.77 & \cellcolor{yellow!40}0.249 & 0.672 \\
MVSNeRF \cite{chen2021mvsnerf}  &&  - & - & - &&17.25 & 0.416 & 0.577 && 18.55 & 0.343 & 0.645 \\
IBRNet \cite{wang2021ibrnet} && \cellcolor{yellow!40}14.93 & \cellcolor{yellow!40}0.625 & \cellcolor{orange!40}0.401 && 18.40 & 0.400 & 0.595 && \cellcolor{yellow!40}21.96 & 0.281 & \cellcolor{yellow!40}0.710 \\
GNT \cite{varma2022attention} && \cellcolor{orange!40}15.99 & \cellcolor{orange!40}0.548 & \cellcolor{yellow!40}0.400 && \cellcolor{orange!40}20.42 & \cellcolor{yellow!40}0.327 & \cellcolor{yellow!40}0.617 && \cellcolor{orange!40}22.47 & \cellcolor{orange!40}0.247 & \cellcolor{orange!40}0.720 \\
Ours && \cellcolor{red!40}17.57& \cellcolor{red!40}0.467 &\cellcolor{red!40} 0.472  && \cellcolor{red!40}21.47 & \cellcolor{orange!40}0.293 & \cellcolor{red!40}0.652 && \cellcolor{red!40}22.74 & \cellcolor{red!40}0.241 & \cellcolor{red!40}0.723 \\ 
\bottomrule
\end{tabular}
}
\end{table}

\begin{table}[tb]
\centering
\caption{Results for generalizable scene rendering on mip-NeRF 360 with few reference views.} 
\label{tab:main_360}
\def\lw{1.5}
\def\ls{0.01}
\resizebox{\linewidth}{!}
{
\begin{tabular}{p{2.3cm}p{\ls cm}p{\lw cm}<{\centering}p{\lw cm}<{\centering}p{\lw cm}<{\centering}p{\ls cm}p{\lw cm}<{\centering}p{\lw cm}<{\centering}p{\lw cm}<{\centering}p{\ls cm}p{\lw cm}<{\centering}p{\lw cm}<{\centering}p{\lw cm}<{\centering}} \toprule
\multirow{2}{*}{Method} &&\multicolumn{3}{c}{1 reference view} && \multicolumn{3}{c}{2 reference views} && \multicolumn{3}{c}{3 reference views}\\
\cline{3-5} \cline{7-9} \cline{11-13} \\ [-8pt]
&&PSNR ($\uparrow$) & LPIPS ($\downarrow$) & SSIM ($\uparrow$) && PSNR ($\uparrow$) & LPIPS ($\downarrow$) & SSIM ($\uparrow$) && PSNR ($\uparrow$) & LPIPS ($\downarrow$) & SSIM ($\uparrow$)\\
\midrule 
MatchNeRF \cite{chen2023explicit} && - & - & - && \cellcolor{orange!40}17.00 & \cellcolor{yellow!40}0.566 & \cellcolor{orange!40}0.392&&\cellcolor{yellow!40} 17.26 & \cellcolor{yellow!40}0.551 & \cellcolor{yellow!40}0.407\\
MVSNeRF \cite{chen2021mvsnerf}  && - & - & - &&14.23 & 0.681 & 0.366&& 14.29 & 0.674 & 0.406\\
IBRNet \cite{wang2021ibrnet} && \cellcolor{orange!40}14.12 & \cellcolor{yellow!40}0.682 & \cellcolor{yellow!40}0.283&& \cellcolor{yellow!40}16.24 & 0.618 & 0.360&& \cellcolor{red!40}17.70 & 0.555 & \cellcolor{orange!40}0.420\\
GNT \cite{varma2022attention} && \cellcolor{yellow!40}13.48 & \cellcolor{orange!40}0.630 & \cellcolor{orange!40}0.314&& 15.21 & \cellcolor{orange!40}0.559 & \cellcolor{yellow!40}0.370 && 15.59 & \cellcolor{orange!40}0.538 & 0.395 \\
Ours  && \cellcolor{red!40}15.20 & \cellcolor{red!40}0.592 & \cellcolor{red!40}0.350 && \cellcolor{red!40}17.05 & \cellcolor{red!40}0.538 & \cellcolor{red!40}0.403 && \cellcolor{orange!40}17.55 & \cellcolor{red!40}0.512 & \cellcolor{red!40}0.430\\ 
\bottomrule
\end{tabular}
}
\end{table}

\begin{table}[tb]
\centering
\def\lw{1.3}
\def\ls{0.06}
\caption{Results on MVImgNet across varying numbers of reference views. `C.' represents the use of calibration before averaging.} 
\label{tab:main_mvimgnet}
\resizebox{\linewidth}{!}
{
\begin{tabular}{p{1.5cm}p{\ls cm}p{\lw cm}<{\centering}p{\lw cm}<{\centering}p{\lw cm}<{\centering}p{\ls cm}p{\lw cm}<{\centering}p{\lw cm}<{\centering}p{\lw cm}<{\centering}p{\ls cm}p{\lw cm}<{\centering}p{\lw cm}<{\centering}p{\lw cm}<{\centering}p{\ls cm}p{\lw cm}<{\centering}p{\lw cm}<{\centering}p{\lw cm}<{\centering}p{\ls cm}p{\lw cm}<{\centering}p{\lw cm}<{\centering}p{\lw cm}<{\centering}} \toprule
\multirow{2}{*}{Method} &&\multicolumn{3}{c}{1 reference view} && \multicolumn{3}{c}{2 reference views} && \multicolumn{3}{c}{3 reference views}&& \multicolumn{3}{c}{4 reference views}&& \multicolumn{3}{c}{5 reference views}\\
\cline{3-5} \cline{7-9} \cline{11-13} \cline{15-17} \cline{19-21} \\ [-8pt]
&&PSNR ($\uparrow$) & LPIPS ($\downarrow$) & SSIM ($\uparrow$) && PSNR ($\uparrow$) & LPIPS ($\downarrow$) & SSIM ($\uparrow$) && PSNR ($\uparrow$) & LPIPS ($\downarrow$) & SSIM ($\uparrow$)&& PSNR ($\uparrow$) & LPIPS ($\downarrow$) & SSIM ($\uparrow$)&& PSNR ($\uparrow$) & LPIPS ($\downarrow$) & SSIM ($\uparrow$)\\
\midrule
IBRNet && 19.14 & 0.458 & 0.595&& 24.38 & 0.266 & 0.818&& 25.53 & 0.203 & 0.858&& 25.99 & 0.190 & 0.867&& 26.12 & 0.188 & 0.867\\
GNT && 22.22 &   0.433 & 0.678 && 26.94 & 0.236 & 0.850&& 27.41 & 0.206 & 0.870&& 27.51 & 0.197 & 0.875&& 27.51 & 0.194 & 0.876\\
Ours \textit{w/o} C. && 23.61 & 0.371 & 0.718&& 26.34 & 0.274 & 0.817&& 27.10 & 0.228 & 0.850 && 27.30 & 0.210 & 0.862 && 27.34 & 0.203 & 0.865 \\
Ours && \cellcolor{gray!50}24.28 & \cellcolor{gray!50}0.334 & \cellcolor{gray!50}0.747&& \cellcolor{gray!50}27.34 & \cellcolor{gray!50}0.215 & \cellcolor{gray!50}0.856&& \cellcolor{gray!50}27.82 & \cellcolor{gray!50}0.190 & \cellcolor{gray!50}0.875&& \cellcolor{gray!50}27.92 & \cellcolor{gray!50}0.181 & \cellcolor{gray!50}0.881&& \cellcolor{gray!50}27.92 & \cellcolor{gray!50}0.179 & \cellcolor{gray!50}0.882\\
\bottomrule
\end{tabular}
}
\end{table}
\textbf{Generalizable rendering.} 
In the generalizable setting, we adopt two training strategies. First, we train the model on multiple datasets as described in~\Cref{sec:exp} and evaluate on LLFF~\cite{mildenhall2019local}, Shiny~\cite{wizadwongsa2021nex} and mip-NeRF 360~\cite{barron2022mip} datasets. In addition, the model is trained and tested on the MVImgNet~\cite{yu2023mvimgnet} for object-centric generalizability.

\underline{(a) LLFF, Shiny, and mip-NeRF 360.} The results for few-reference view scenarios on these datasets are shown in~\Cref{tab:main_few_llff,,tab:main_shiny,,tab:main_360}, respectively. Methods like MatchNeRF~\cite{chen2023explicit}, MVSNeRF~\cite{yao2018mvsnet}, and GeoNeRF~\cite{johari2022geonerf} require at least two reference views. On the LLFF dataset, all methods experience a performance decline as the number of views decreases. CaesarNeRF, however, consistently outperforms others across varying reference view numbers, with the performance gap becoming more significant with fewer views. For example, with 3 views, while IBRNet~\cite{wang2021ibrnet} and GNT~\cite{varma2022attention} have comparable PSNRs, CaesarNeRF demonstrates a more substantial lead in LPIPS and SSIM metrics.

Similar patterns are observed on the Shiny~\cite{wizadwongsa2021nex} and mip-NeRF 360~\cite{barron2022mip} datasets. We apply the highest-performing methods from the LLFF evaluations and report the results for those that produce satisfactory outcomes with few reference views. CaesarNeRF maintains superior performance throughout. Notably, for complex datasets like mip-NeRF 360~\cite{barron2022mip}, which have sparse camera inputs, the quality of rendered images generally decreases with fewer available reference views. Nonetheless, CaesarNeRF shows the most robust performance compared to the other methods.

\begin{table}[tb]
\centering
\caption{Results of per-scene optimization on LLFF, in comparison with state-of-the-art methods.} 
\label{tab:main_per_scene_combined}
\def\lw{2}
\def\lk{2.4}
\def\ls{0.06}
\resizebox{.85\linewidth}{!}
{
\begin{tabular}{p{\lw cm}p{\ls cm}p{\lw cm}<{\centering}p{\lw cm}<{\centering}p{\lw cm}<{\centering}p{\lw cm}<{\centering}p{\lw cm}<{\centering}p{\lw cm}<{\centering}p{\lw cm}<{\centering}} \toprule

\multirow{1}{*}{Method} && LLFF \cite{mildenhall2019local}  & NeRF \cite{mildenhall2021nerf} & NeX \cite{wizadwongsa2021nex} & GNT \cite{varma2022attention} & Ours\\
\midrule
 PSNR ($\uparrow$)&&  23.27 & 26.50 & \cellcolor{orange!40}27.26 & \cellcolor{yellow!40}27.24 & \cellcolor{red!40}27.64\\
 LPIPS ($\downarrow$)&& 0.212 & 0.250 & \cellcolor{yellow!40}0.179  & \cellcolor{orange!40}0.087 & \cellcolor{red!40}0.081\\
SSIM ($\uparrow$)  && 0.798 & 0.811& \cellcolor{red!40}0.904 &  \cellcolor{yellow!40}0.889& \cellcolor{red!40}0.904 \\
 \bottomrule
\end{tabular}
}
\end{table}



\begin{table}[t]
\centering
\caption{Results on LLFF for few-shot generalization after adapting Caesar to other baselines.} 
\label{tab:main_generalizable_model}
\def\lw{1.5}
\def\lk{2.9}
\def\ls{0.0}
\resizebox{0.95\linewidth}{!}
{
\begin{tabular}{p{2.cm}p{\ls cm}p{\lw cm}<{\centering}p{\lw cm}<{\centering}p{\lw cm}<{\centering}p{\ls cm}p{\lw cm}<{\centering}p{\lw cm}<{\centering}p{\lw cm}<{\centering}p{\ls cm}p{\lw cm}<{\centering}p{\lw cm}<{\centering}p{\lw cm}<{\centering}} \toprule
\multirow{2}{*}{Method} &&\multicolumn{3}{c}{1 reference view} && \multicolumn{3}{c}{2 reference views} && \multicolumn{3}{c}{3 reference views}\\
\cline{3-5} \cline{7-9} \cline{11-13} \\ [-8pt]
&&PSNR ($\uparrow$) & SSIM ($\uparrow$) & LPIPS ($\downarrow$) &&PSNR ($\uparrow$) & SSIM ($\uparrow$) & LPIPS ($\downarrow$) &&PSNR ($\uparrow$) & SSIM ($\uparrow$) & LPIPS ($\downarrow$) \\
\midrule 
\multirow{1}{*}{MatchNeRF \cite{chen2023explicit}} && - & - & - &&20.59 & 0.775& 0.276  && 22.43 & 0.805& 0.244  \\
\quad + Caesar && - & - & - &&  21.55 & 0.782& 0.268  &&  22.98 & 0.824& 0.242  \\
\midrule
\multirow{1}{*}{IBRNet \cite{wang2021ibrnet}} && 16.85 & 0.507  & 0.542 & & 21.25 & 0.685 & 0.333  && 23.00 & 0.752 & 0.262   \\
\quad + Caesar& & 17.76 & 0.543 & 0.500 && 22.39 & 0.740 & 0.275 && 23.67 & 0.772 & 0.242   \\
\bottomrule
\end{tabular}
}
\end{table}

\begin{table}[tb]
\centering
\caption{Ablations on the  semantic representation length $\bm{R}$, sequential refinement (Seq.) and calibration (Cali.). `Ext.' denotes the extension of per-pixel feature to a length of 64 in GNT.} 
\label{tab:latent_ablation}
\def\lw{1.6}
\def\lk{2.4}
\def\ls{0.0}
\resizebox{0.7\linewidth}{!}
{
\begin{tabular}{p{\lw cm}<{\centering}p{\lw cm}<{\centering}p{\lw cm}<{\centering}p{\ls cm}p{\lw cm}<{\centering}p{\lw cm}<{\centering}p{\lw cm}<{\centering}p{\lw cm}<{\centering}p{\lw cm}<{\centering}} \toprule

\multicolumn{3}{c}{Model Variations} && \multirow{2}{*}{PSNR ($\uparrow$)}  & \multirow{2}{*}{LPIPS ($\downarrow$)} & \multirow{2}{*}{SSIM ($\uparrow$)}  \\
\cline{1-3} \\ [-8pt]
 $\vb*R$ len. & Seq. & Cali. \\ 
\midrule
 \multicolumn{3}{c}{(Baseline GNT)} && 20.93 & 0.185 & 0.731 \\
 \midrule
 Ext. && && 20.85 & 0.173 & 0.735\\
+32 && && 21.43 & 0.152 & 0.763 \\
+64 &&  &&21.49 & 0.149 & 0.766\\
+96 &&  &&21.46 & 0.150 & 0.766\\
+128 &&  &&21.49 & 0.147 & 0.763\\
\midrule
+ 96 &\cmark &   && 21.53 & 0.146 & 0.770\\
+ 96 && \cmark &&21.51 & 0.147 & 0.769\\
+ 96 & \cmark & \cmark&& \cellcolor{gray!50}21.67 & \cellcolor{gray!50}0.139 & \cellcolor{gray!50}0.781\\
\bottomrule
\end{tabular}
}
\end{table}

\begin{figure*}[t]
    \begin{center}
  \def\fw{0.1952}
    \resizebox{\linewidth}{!}{
    \begin{tabular}{p{0.2 \linewidth}<{\centering}p{0.2\linewidth}<{\centering}p{0.2\linewidth}<{\centering}p{0.2\linewidth}<{\centering}p{0.2\linewidth}<{\centering}}
   IBRNet~\cite{wang2021ibrnet} & GPNR~\cite{suhail2022generalizable} & NeuRay~\cite{liu2022neural} & CaesarNeRF & Ground-truth \\
   \end{tabular}}
    \includegraphics[width=\fw\linewidth]{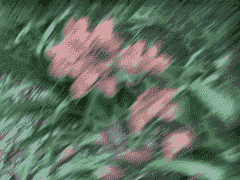} \hfill  \includegraphics[width=\fw\linewidth]{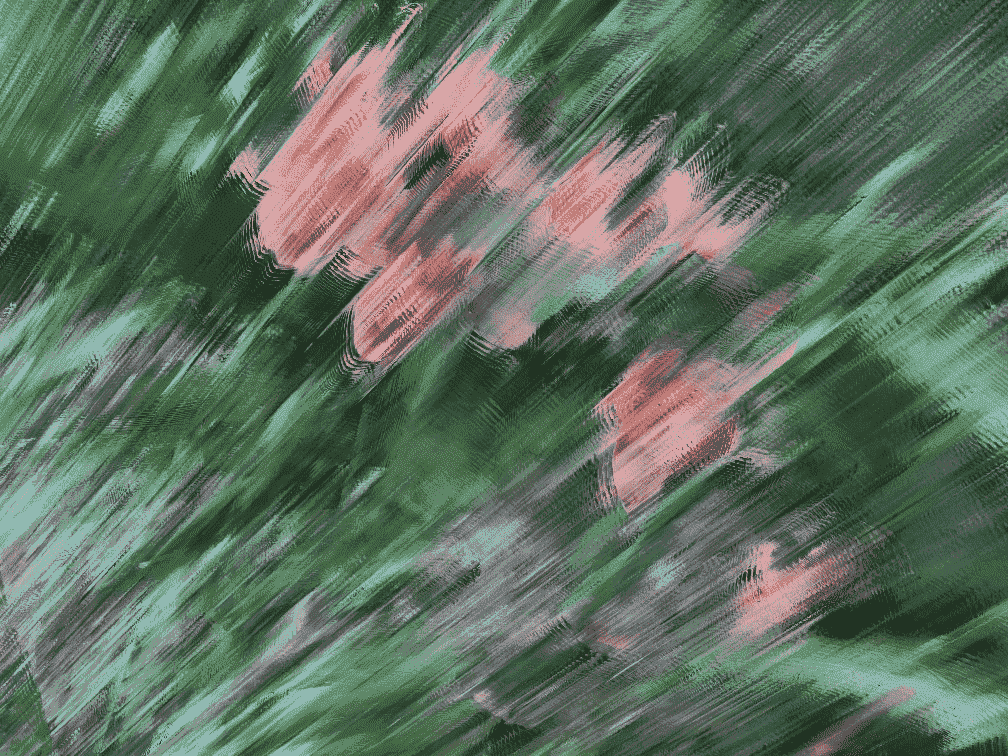} \hfill \includegraphics[width=\fw\linewidth]{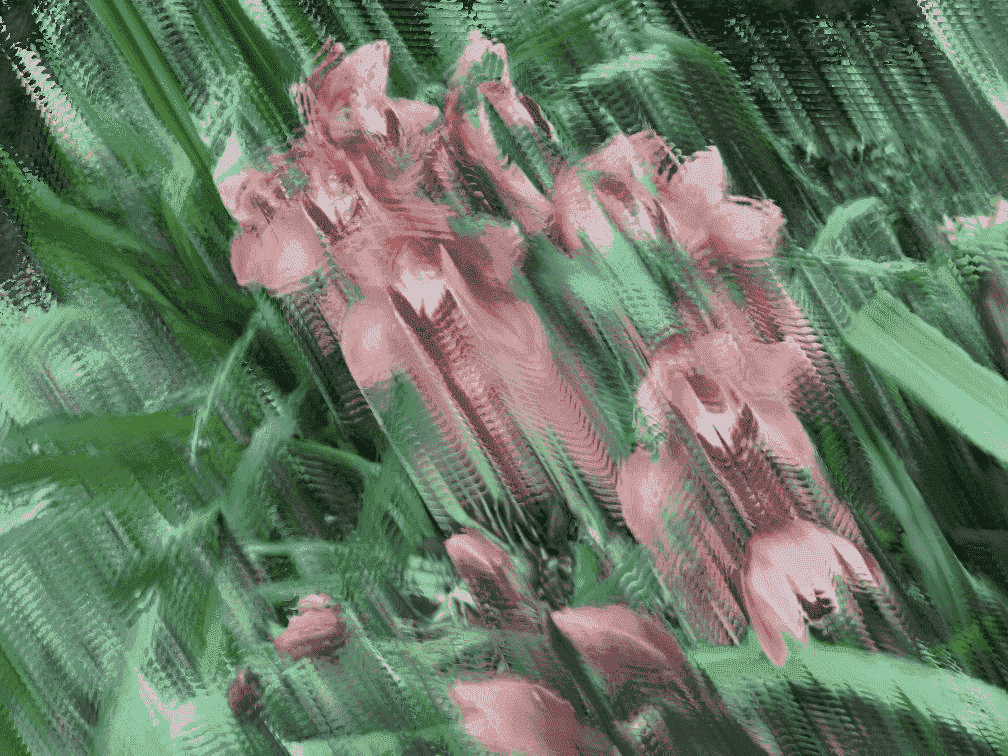} \hfill  \includegraphics[width=\fw\linewidth]{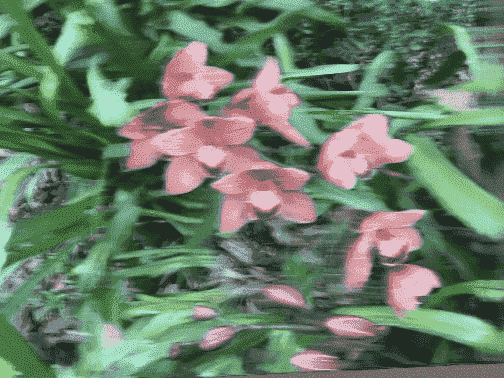} \hfill  \includegraphics[width=\fw\linewidth]{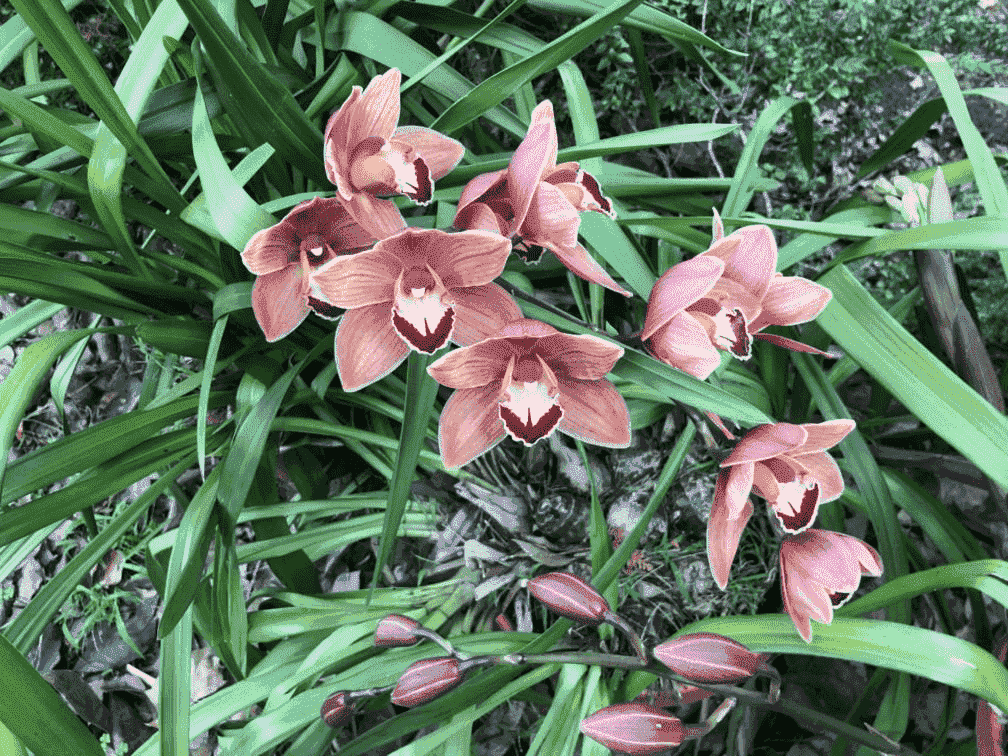} \\
    \includegraphics[width=\fw\linewidth]{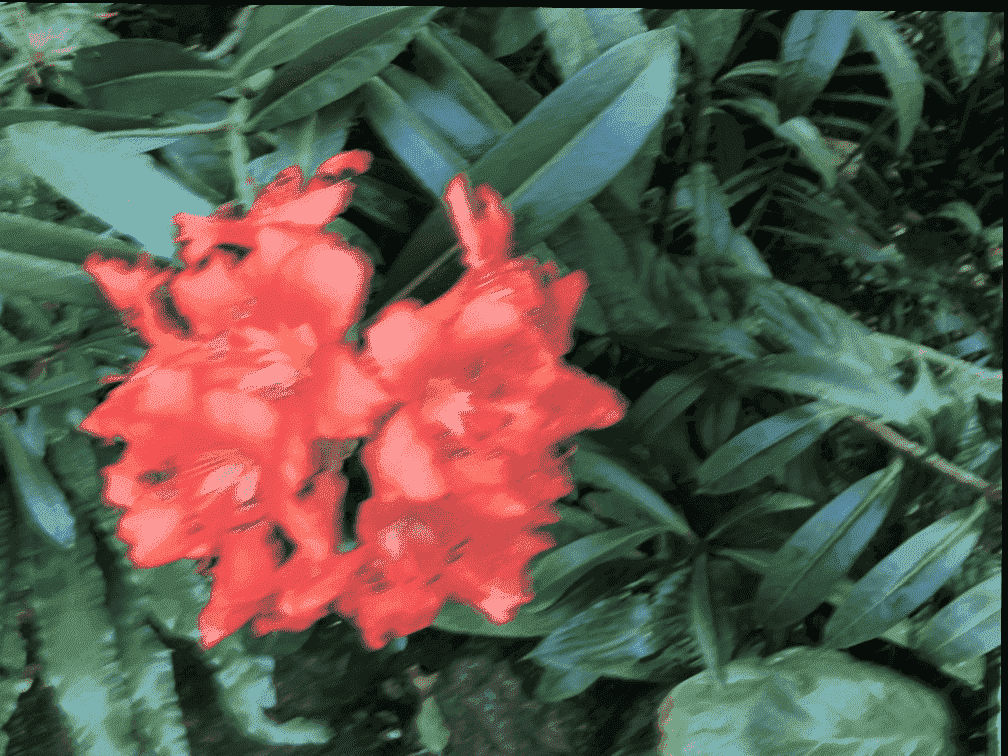} \hfill  \includegraphics[width=\fw\linewidth]{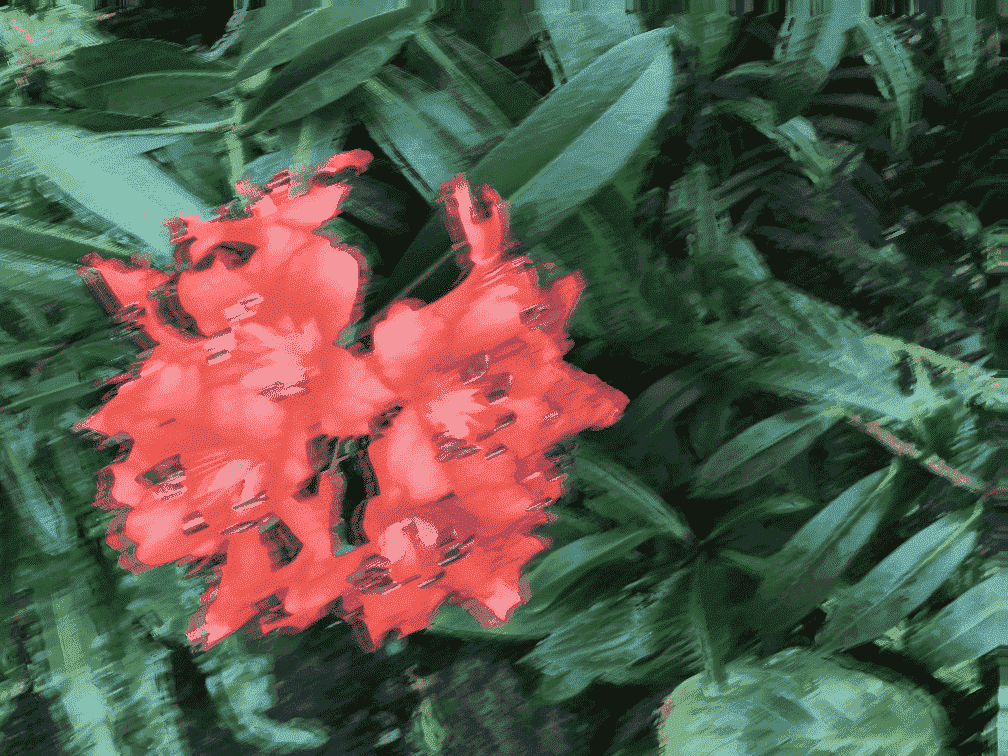} \hfill  \includegraphics[width=\fw\linewidth]{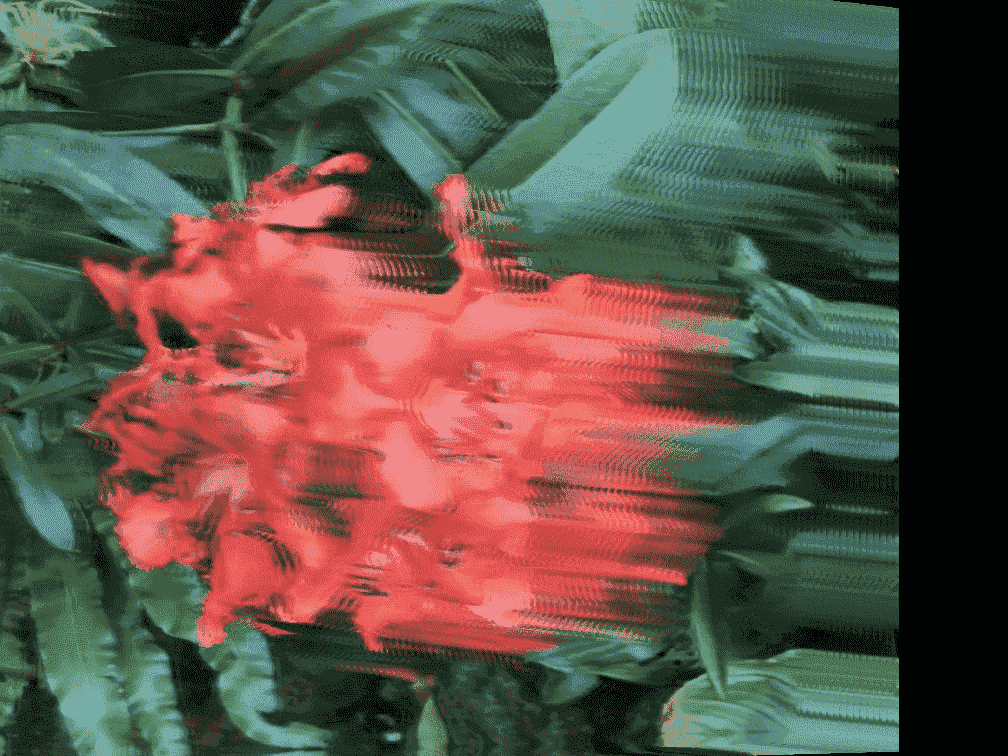} \hfill  \includegraphics[width=\fw\linewidth]{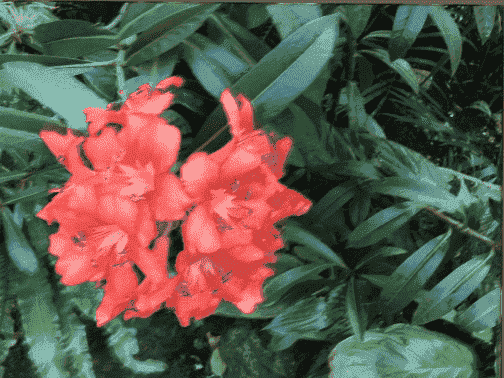} \hfill  \includegraphics[width=\fw\linewidth]{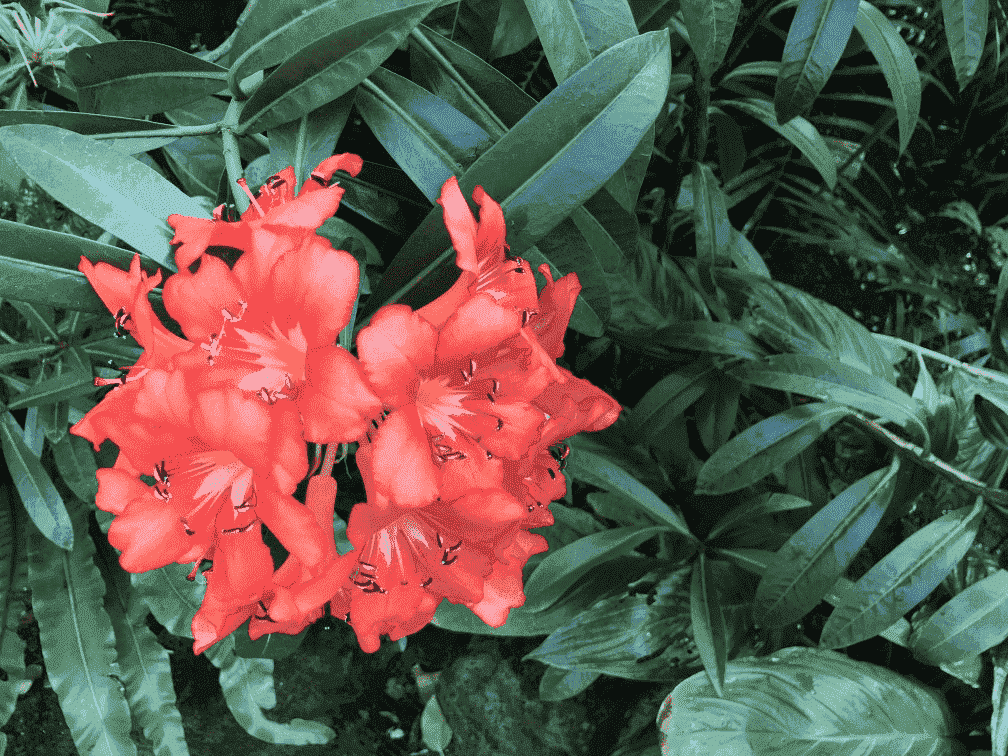} \\
    \small{(a) Using one image as reference view.}\\
    \includegraphics[width=\fw\linewidth]{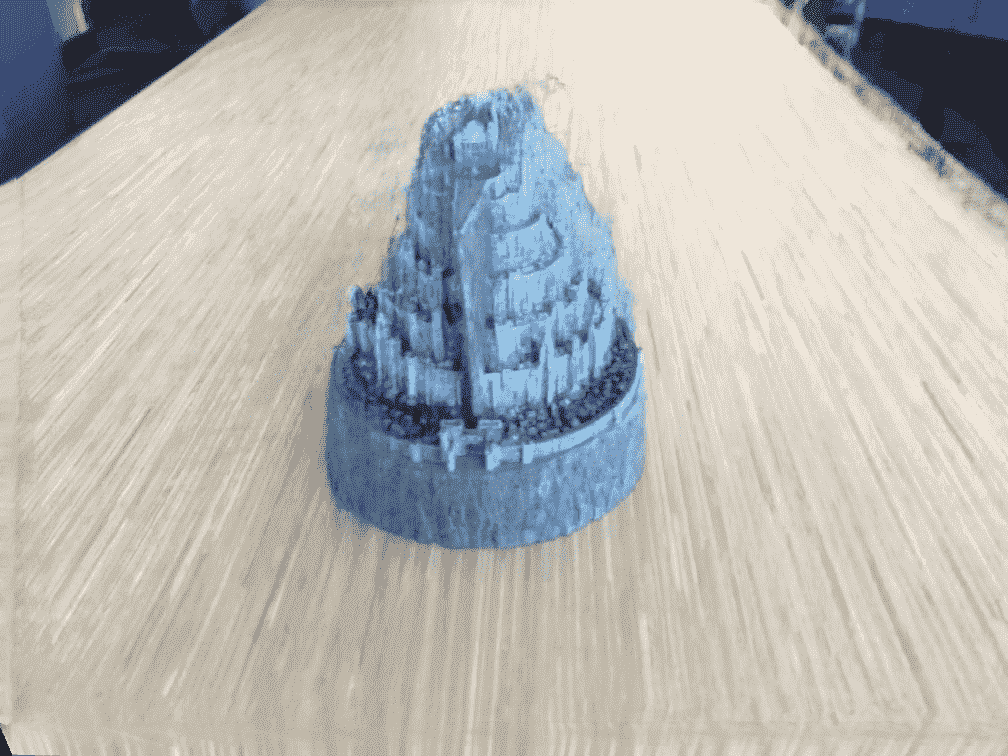} \hfill  \includegraphics[width=\fw\linewidth]{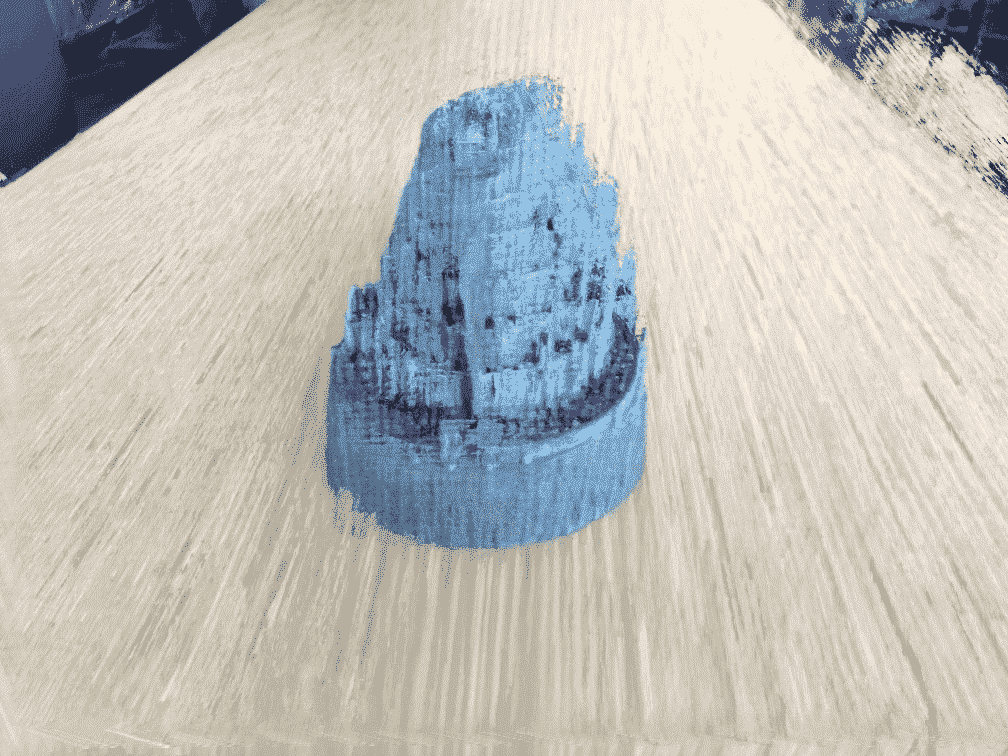} \hfill  \includegraphics[width=\fw\linewidth]{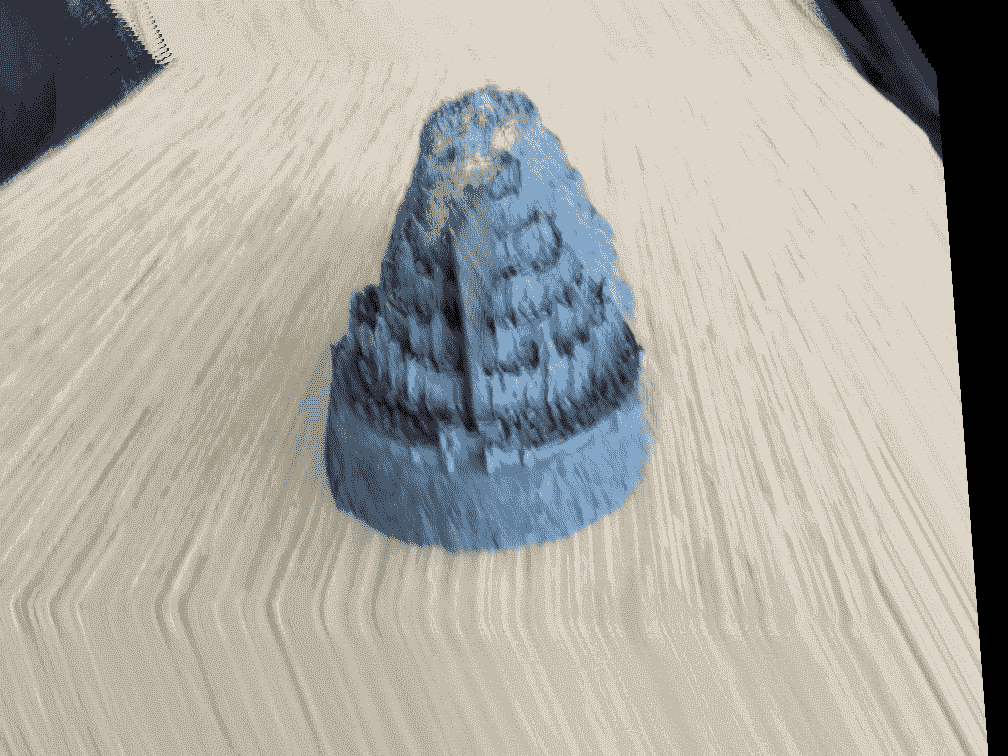} \hfill  \includegraphics[width=\fw\linewidth]{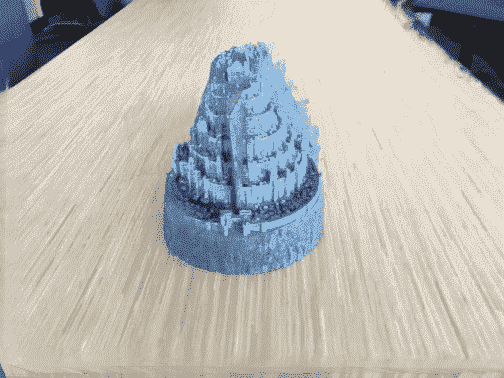} \hfill  \includegraphics[width=\fw\linewidth]{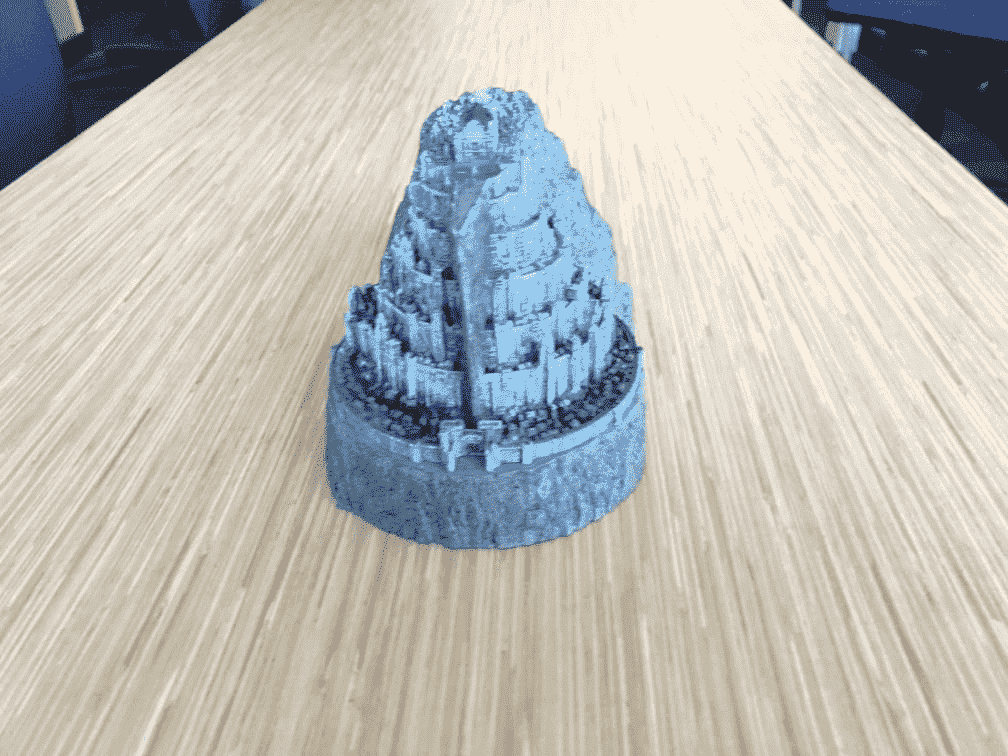} \\
    \includegraphics[width=\fw\linewidth]{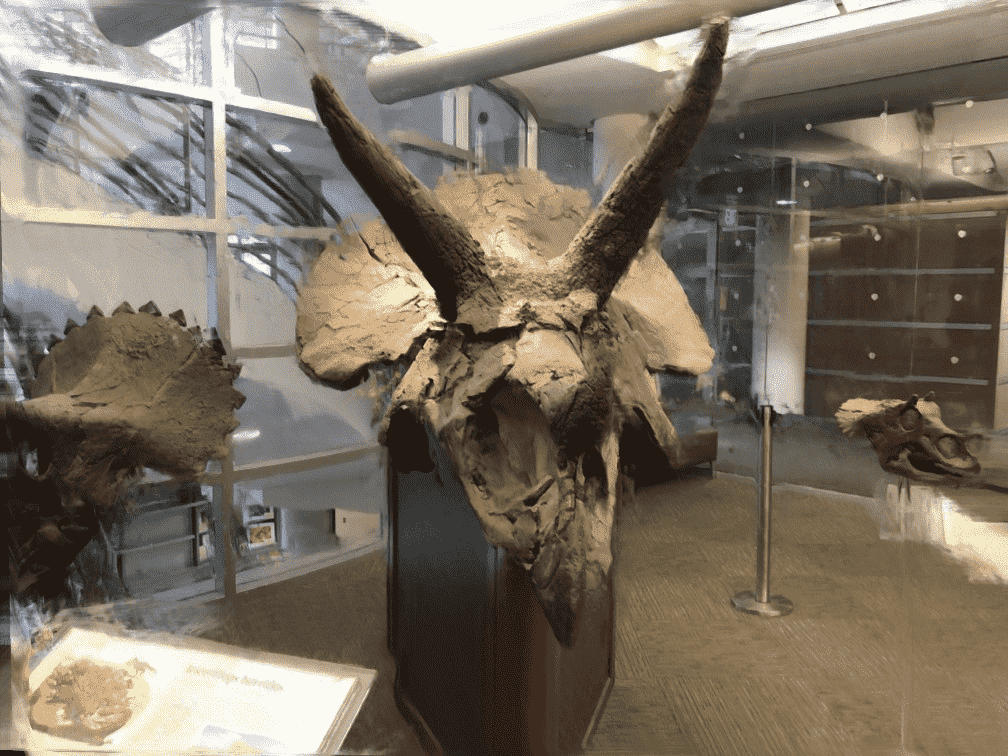} \hfill  \includegraphics[width=\fw\linewidth]{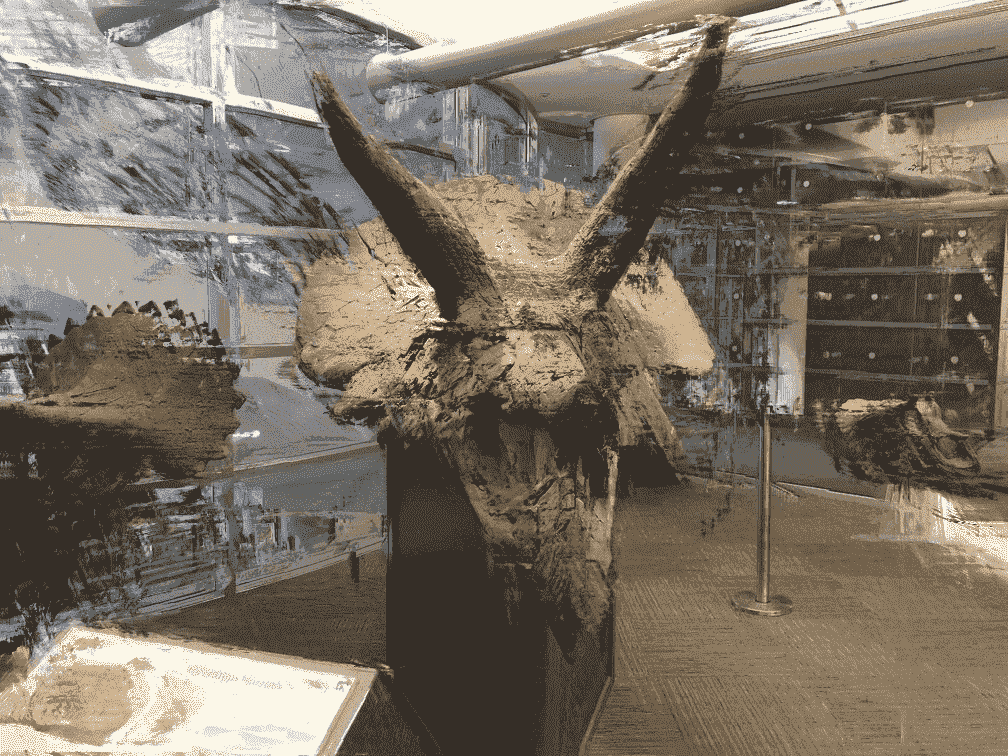} \hfill  \includegraphics[width=\fw\linewidth]{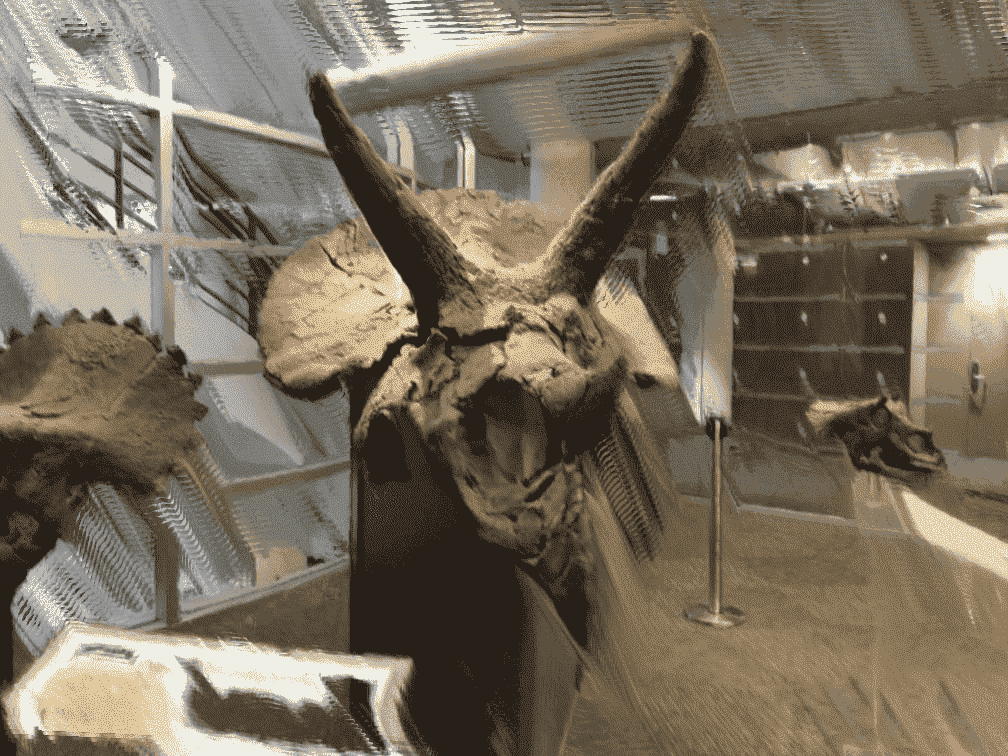} \hfill  \includegraphics[width=\fw\linewidth]{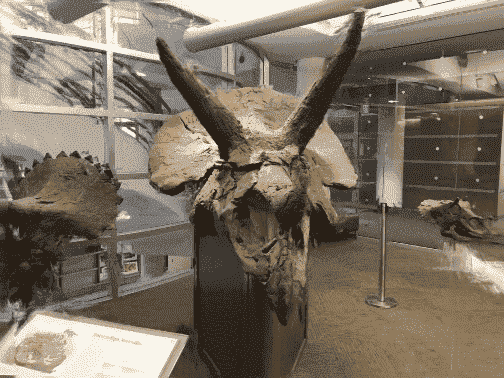} \hfill  \includegraphics[width=\fw\linewidth]{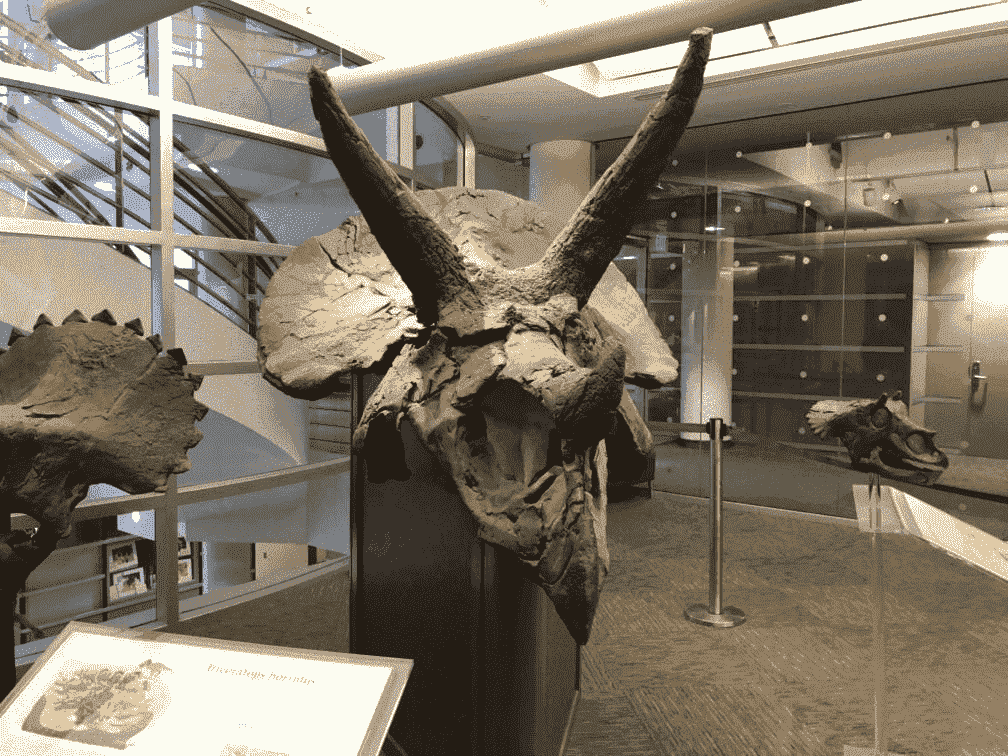} \\
    \small{(b) Using two images as reference view.}
    \caption{Comparative visualization of our proposed method against other state-of-the-art methods.}
    \label{fig:vis}
    \end{center}
\end{figure*}

\begin{figure*}[t]
    \centering
    \def\lw{3.5}
    \resizebox{\linewidth}{!}{
        \begin{tabular}{p{\lw cm}<{\centering}p{\lw cm}<{\centering}p{\lw cm}<{\centering}p{\lw cm}<{\centering}p{\lw cm}<{\centering}p{\lw cm}<{\centering}}
         Target Image & GNT & CaesarNeRF & Target Image & GNT & CaesarNeRF \\
    \end{tabular}}
    \includegraphics[width=\linewidth]{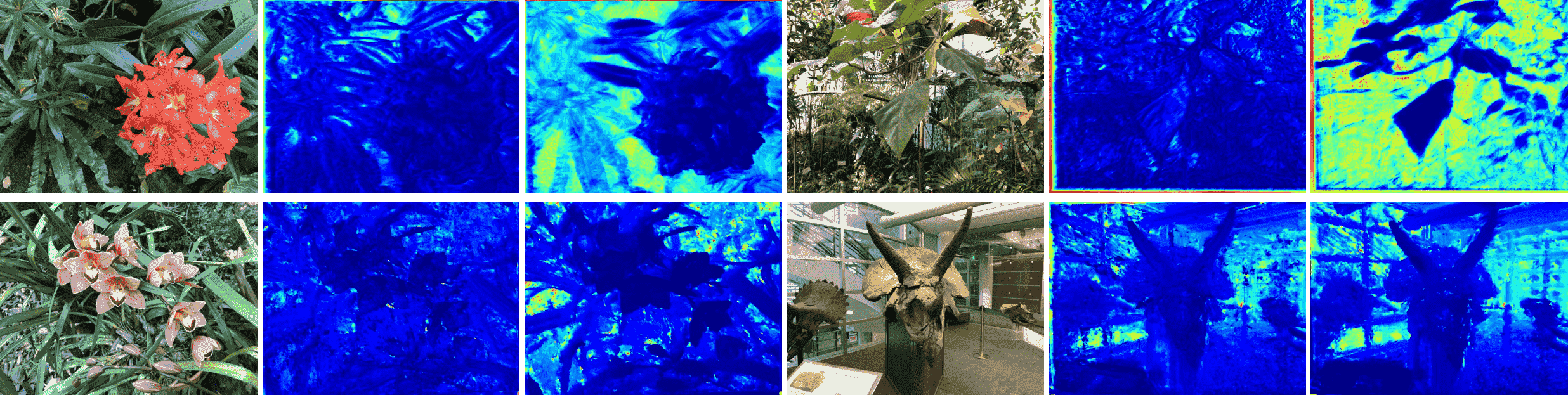}
    \caption{Depth estimation prediction using one reference view (first row) and two reference views (second row) as input from LLFF comparing CaesarNeRF with GNT.}
    \label{fig:depth}
\end{figure*}
\underline{(b) MVImgNet.}
We extend our comparison of CaesarNeRF with GNT~\cite{varma2022attention} and IBRNet~\cite{wang2021ibrnet} on the MVImgNet dataset, focusing on object-centric scenes, as shown in~\Cref{tab:main_mvimgnet}. We examine a variant of CaesarNeRF where semantic calibration is substituted with simple feature averaging from multiple frames. While the performance of all methods improves with more views, CaesarNeRF consistently outperforms GNT and IBRNet. Notably, CaesarNeRF with feature averaging surpasses GNT in 1-view case but lags with additional views, implying that the absence of calibration lead to ambiguities when rendering from multiple views.

\textbf{Per-scene optimization.}
Beyond the multi-scene generalizable setting, we demonstrate per-scene optimization results in~\Cref{tab:main_per_scene_combined}. We calculate the average performance over 8 categories from the LLFF dataset~\cite{mildenhall2019local}. CaesarNeRF consistently outperforms nearly all state-of-the-art methods in the comparison, across all three metrics, showing a significant improvement over our baseline method, GNT~\cite{varma2022attention}. 

\textbf{Adaptability.} To test the adaptability of our Caesar pipeline, we apply it to two other state-of-the-art methods that use \textit{view transformers}, namely MatchNeRF~\cite{chen2023explicit} and IBRNet~\cite{wang2021ibrnet}. We demonstrate in~\Cref{tab:main_generalizable_model} that our enhancements in scene-level semantic understanding significantly boost the performance of these methods across all metrics. This indicates that the Caesar framework is not only beneficial in our CaesarNeRF, which is based on GNT~\cite{varma2022attention}, but can also be a versatile addition to other NeRF pipelines with view transformers to aggregate different input views.

\textbf{Ablation analysis.} We conduct ablation studies on the ``orchid'' scene from the LLFF dataset, with findings detailed in~\Cref{tab:latent_ablation}. Testing variations in representation and the impact of the sequential refinement and calibration modules, we find that increasing the latent size in GNT yields marginal benefits. However, incorporating even a modest semantic representation size distinctly improves results. The length of the semantic representation has a minimal impact on quality. Our ablation studies indicate that while sequential refinement and calibration each offer slight performance gains, their combined effect is most significant. In a single-scene context, semantic information is effectively embedded within the representation, making the benefits of individual modules subtler. Together, however, they provide a framework where sequential refinement can leverage calibrated features for deeper insights.

\textbf{Visualizations.} We present our visualization results in~\Cref{fig:vis}, where we compare our method with others using one or two views from the LLFF dataset. Additional visual comparisons are provided in the supplementary materials. These visualizations highlight that in scenarios with few views, our method significantly surpasses other generalizable NeRF models, particularly excelling when only a single view is available. CaesarNeRF demonstrates rendering with sharper boundaries and more distinct objects.

\textbf{Depth estimation.} We extend our evaluation to depth prediction within the LLFF~\cite{mildenhall2019local} dataset, focusing on challenges presented by few reference views, such as scenarios with just one or two images. In a comparison between CaesarNeRF and GNT~\cite{varma2022attention}, we observe in~\Cref{fig:depth} that GNT struggles to accurately capture the relative positions of objects when reference images are sparse. For instance, with only a single view of a flower, CaesarNeRF precisely indicates the flower's proximity to the camera compared to the leaves in the background, a distinction that GNT fails to make. Furthermore, the depth estimations provided by CaesarNeRF are consistently more reliable. In the horn example involving two views, CaesarNeRF offers better boundary delineation, showing particular strength in handling reflective surfaces such as grass in the background.

\begin{figure*}[!t]
    \begin{center}
    \def\fw{0.19}
    \footnotesize
     \setlength{\tabcolsep}{6pt}
     \resizebox{\linewidth}{!}{
         \begin{tabular}{c|cc|cc@{}}
         Source Image & Smallest Distance & $2^{nd}$ Smallest Distance & Largest Distance & $2^{nd}$ Largest Distance \\
         \includegraphics[width=\fw\linewidth]{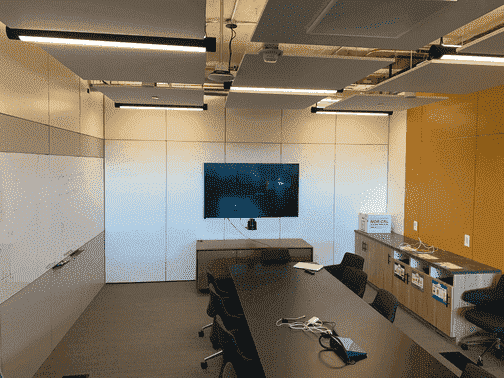} &  \includegraphics[width=\fw\linewidth]{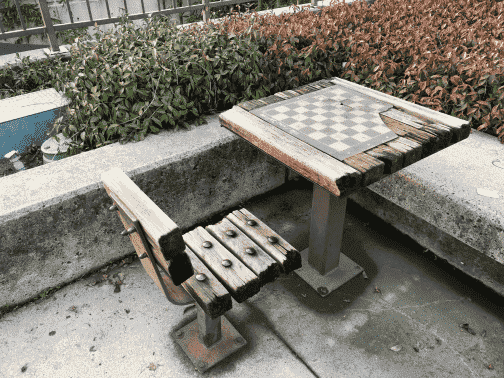} &  \includegraphics[width=\fw\linewidth]{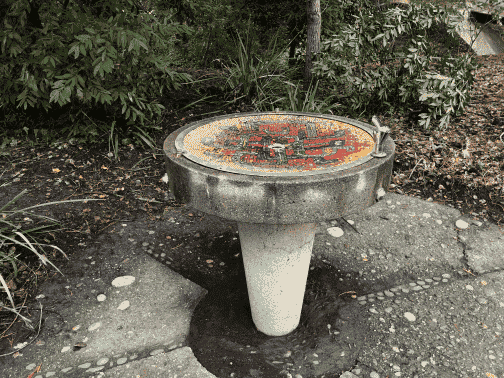} &  \includegraphics[width=\fw\linewidth]{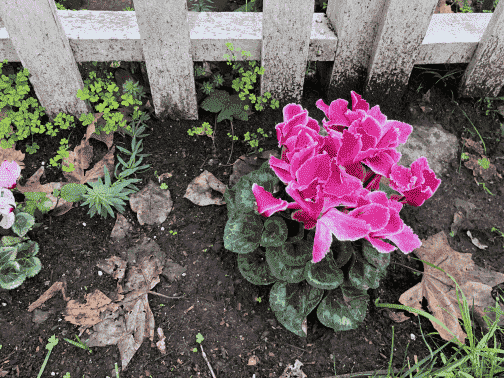} &  \includegraphics[width=\fw\linewidth]{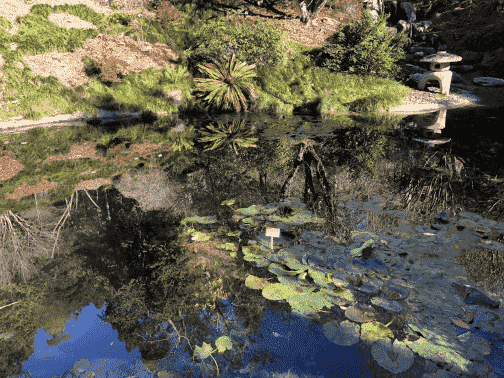} \\
         fortress & chesstable (0.45) & colorfountain (0.50) & fenceflower (1.91) & pond (1.80)  \\
         \includegraphics[width=\fw\linewidth]{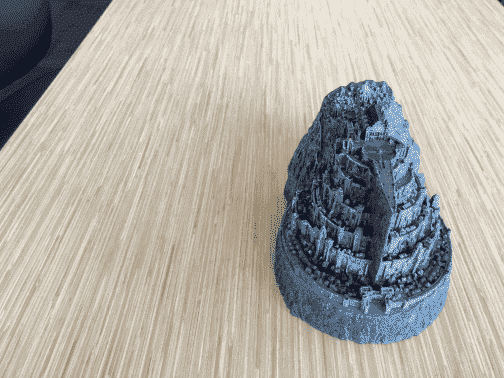} &  \includegraphics[width=\fw\linewidth]{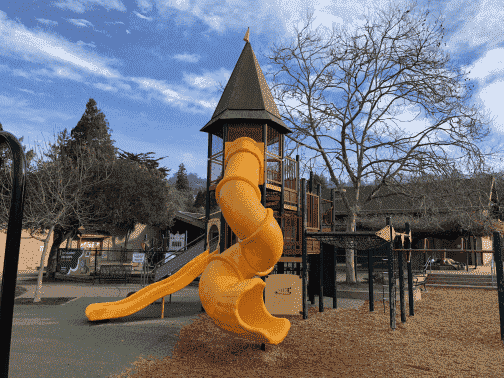} &  \includegraphics[width=\fw\linewidth]{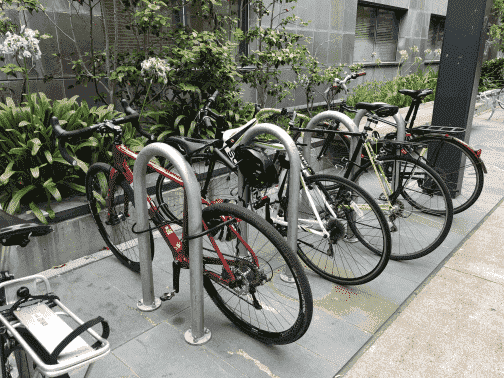} &  \includegraphics[width=\fw\linewidth]{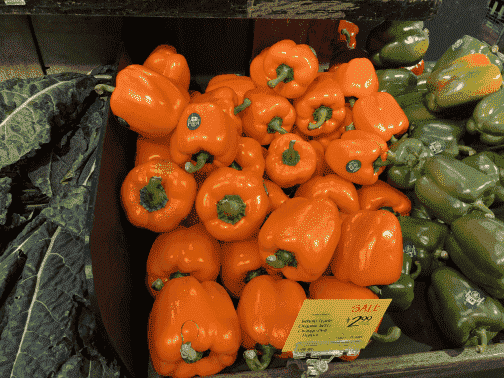} &  \includegraphics[width=\fw\linewidth]{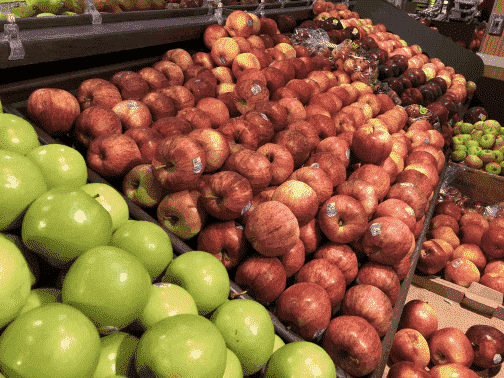} \\
         fern & playground (0.46) & bikes (0.50) & peppers (2.02) & apples (2.01)  \\
         \end{tabular}        
         }
    \caption{Largest and smallest distances for two examples from LLFF test split when matching with training scenes. Numbers ($\times 10^{-2}$) denote the L-2 distance to the source image within the semantic feature space.}
    \label{fig:semantic}
    \end{center}%
\end{figure*}

\textbf{Semantic analysis for $\Tilde{\vb*{S}}$.} To assess whether the calibrated semantic representation $\Tilde{\vb*{S}}$ truly captures semantic details of the scene, we analyze the highest and lowest response based on the L-2 distance between features from two different scenes, interpreting smaller distances as greater similarity. We consider the first image from each category in the LLFF dataset as a reference and extract the scene-level representation for the ten closest views to this reference image using CaesarNeRF.

In~\Cref{fig:semantic}, we present two examples, ``room'' and ``fortress'', from the LLFF dataset, representing the first image of these categories alongside images corresponding to the top-2 highest and lowest responses. This analysis reveals that the scene-level representation predominantly emphasizes structural information and objects of similar categories. For instance, with the source image ``room'', the highest responses correlate with images featuring table-like structures, indicating an object-centric focus. In contrast, the lowest responses include images of flowers or ponds found in open areas, which diverge significantly from the structural and object content in the source images.

\begin{figure}[t]
    \centering
    \resizebox{\linewidth}{!}
    {
    \begin{tabular}{cccc}
    \includegraphics[width=0.23 \linewidth]{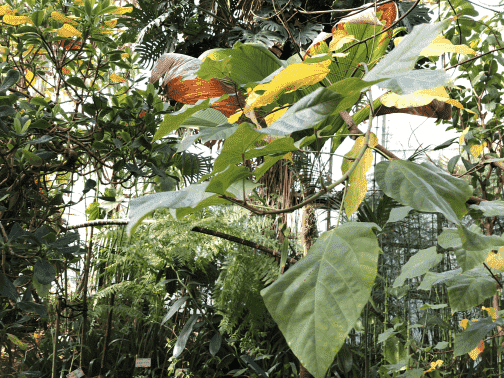} &\includegraphics[width=0.23 \linewidth]{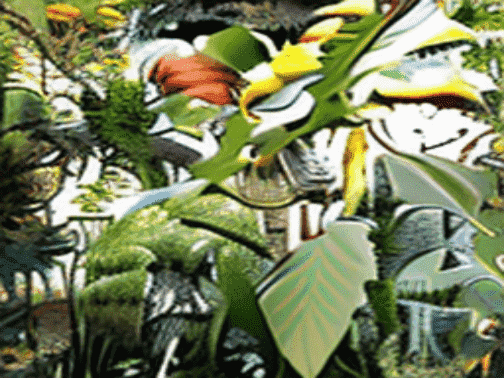} &  \includegraphics[width=0.23 \linewidth]{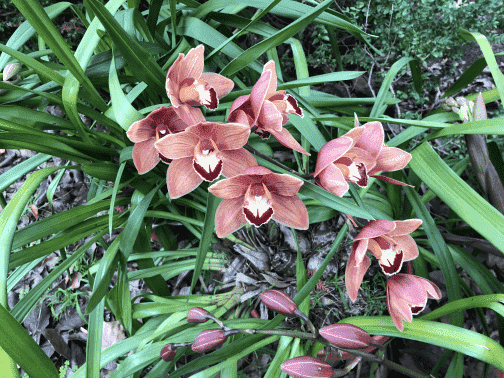} &  \includegraphics[width=0.23 \linewidth]{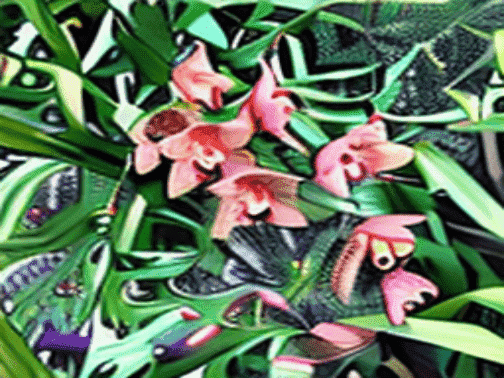} \\
     \multicolumn{2}{c}{(a) leaves} & \multicolumn{2}{c}{(b) orchid} \\
    \end{tabular}
    }
    \caption{Synthetic results for two examples from LLFF~\cite{mildenhall2019local}, ``leaves'' and ``orchid'', using zero 1-to-3~\cite{liu2023zero} with one reference image as input and 2 degrees of vertical shift. The left image of each pair is the input, and the right one is the output of zero 1-to-3~\cite{liu2023zero}.}
    \label{fig:zero123}
\end{figure}

\textbf{Comparison with generative methods.} Single-view scenarios are often addressed by generative methods~\cite{liu2023zero,liu2023syncdreamer,liu2023deceptive} that employ diffusion models~\cite{ho2020denoising,rombach2022high}.  While these models can produce reasonable results for object-centric renderings, they struggle with scene-level renderings from novel viewpoints. We show two LLFF examples in \Cref{fig:zero123} using zero 1-to-3~\cite{liu2023zero}, where the left image is the input, and the right one is the output. Images rendered with zero 1-to-3~\cite{liu2023zero} suffer from style difference. Unlike NeRF-based approaches that reconstruct images from observed pixels, generative models synthesize an entire image from its semantic representation but may not maintain the style.

\section{Conclusion and limitation}\label{sec:con}

In this paper, we introduce CaesarNeRF, a few-shot and generalizable NeRF pipeline  that combines scene-level semantic with per-pixel feature representations, aiding in rendering from novel camera positions with limited reference views. We calibrate the semantic representations across different input views and employ a sequential refinement network to offer distinct semantic representations at various levels. Our method has been extensively evaluated on a broad range of datasets, exhibiting state-of-the-art performance in both generalizable and single-scene settings.

\textbf{Limitations and potential negative impact.} Instead of a generative approach, CaesarNeRF relies on NeRF's scheme, using input images to render the target view. This approach restricts its ability to render parts of the scene that are not present in the reference images. CaesarNeRF may create fake images for authentication.
\bibliographystyle{splncs04}
\bibliography{main}

\end{document}